\documentclass[sigconf, 9pt]{acmart}

\copyrightyear{2021}
\acmYear{2021}
\setcopyright{acmcopyright}\acmConference[SIGMOD '21]{Proceedings of the 2021
International Conference on Management of Data}{June 20--25, 2021}{Virtual Event,
China}
\acmBooktitle{Proceedings of the 2021 International Conference on Management of
Data (SIGMOD '21), June 20--25, 2021, Virtual Event, China}
\acmPrice{15.00}
\acmDOI{10.1145/3448016.3457321}
\acmISBN{978-1-4503-8343-1/21/06}

\usepackage{mathtools}      

\usepackage{xspace}
\newcommand{\name}{ARM-Net\xspace}
\newcommand{\fullname}{Adaptive Relation Modeling Network\xspace}
\newcommand{\nameplus}{ARM-Net+\xspace}
\newcommand{\mname}{ARM\xspace}
\newcommand{\fullmname}{Adaptive Relation Modeling Module\xspace}
\newcommand{\framework}{ARMOR\xspace}

\newcommand{\frappe}{Frappe\xspace}
\newcommand{\movielens}{MovieLens\xspace}
\newcommand{\avazu}{Avazu\xspace}
\newcommand{\criteo}{Criteo\xspace}
\newcommand{\diabetes}{Diabetes130\xspace}

\usepackage{amsopn}
\DeclareMathOperator{\entmax}{\alpha-entmax}

\newcommand{\diag}{\mathop{\mathrm{diag}}}
\newcommand{\logloss}{\mathop{\mathrm{Logloss}}}

\usepackage{color}

\newcommand{\highlight}[1]{\textbf{#1}}
\newcommand{\revise}[1]{{\color{black}#1}}
\newcommand{\update}[1]{{\color{black}#1}}

\newcommand{\powerpoint}[1]{\textbf{#1}}

\usepackage{enumitem}
\newcommand{\squishlist}
{
	\begin{list}{$\bullet$}
		{
			\setlength{\itemsep}{0pt}
			\setlength{\parsep}{3pt}
			\setlength{\topsep}{3pt}
			\setlength{\partopsep}{0pt}
			\setlength{\leftmargin}{1.5em}
			\setlength{\labelwidth}{1em}
			\setlength{\labelsep}{0.5em}
		}
	}
	
	\newcommand{\squishend}
	{
	\end{list}
}

\usepackage{multirow}

\usepackage{caption}
\usepackage{subcaption}

\newsavebox{\measurebox}

\usepackage{multicol}

\newcommand{\unsim}{\mathord{\sim}}

\begin{document}
\fancyhead{}


\title[\fullname for Structured Data]{\name: \fullname \\ for Structured Data}



\author{Shaofeng Cai}
\affiliation{%
  \institution{National University of Singapore}
}
\email{shaofeng@comp.nus.edu.sg}

\author{Kaiping Zheng}
\affiliation{%
  \institution{National University of Singapore}
}
\email{kaiping@comp.nus.edu.sg}

\author{Gang Chen}
\affiliation{%
  \institution{Zhejiang University}
}
\email{cg@zju.edu.cn}

\author{H. V. Jagadish}
\affiliation{%
  \institution{University of Michigan}
}
\email{jag@umich.edu}

\author{Beng Chin Ooi}
\affiliation{%
  \institution{National University of Singapore}
}
\email{ooibc@comp.nus.edu.sg}

\author{Meihui Zhang}
\affiliation{%
  \institution{Beijing Institute of Technology}
}
\email{meihui\_zhang@bit.edu.cn}



\begin{abstract}
 
Relational databases are the $de facto$ standard for storing and querying structured data, and extracting insights from structured data requires advanced analytics.
Deep neural networks (DNNs) have achieved super-human prediction performance in particular data types, e.g., images.
However, existing DNNs may not produce meaningful results when applied to structured data. The reason is that there are correlations and dependencies across combinations of attribute values in a table, \update{and these do not follow simple additive patterns that can be easily mimicked by a DNN.}
The number of possible such ``cross features'' is combinatorial, making them computationally prohibitive to model.
\update{Furthermore, the deployment of learning models in real-world applications has also highlighted the need for interpretability, especially for high-stakes applications, which remains another issue of concern to DNNs.}

\update{In this paper, we present \name, an adaptive relation modeling network tailored for structured data, and a lightweight framework \framework based on \name for relational data analytics.}
The key idea is to \revise{model feature interactions with cross features selectively and dynamically}, by first transforming the input features into exponential space, and then determining the \textit{interaction order} and \textit{interaction weights} adaptively for each cross feature.
We propose a novel sparse attention mechanism to dynamically generate the interaction weights given the input tuple, \update{so that we can explicitly model cross features of arbitrary orders with noisy features filtered selectively.}
Then during model inference, \update{\name can specify the cross features being used for each prediction for higher accuracy and better interpretability.
Our extensive experiments on real-world datasets demonstrate that \name consistently outperforms existing models and provides more interpretable predictions for data-driven decision making.}

\end{abstract}

\ccsdesc[500]{Computing methodologies~Neural networks}
\ccsdesc[100]{Mathematics of computing}
\ccsdesc[100]{Applied computing}

\keywords{Neural Networks; Structured data; Feature interaction; Multi-Head Gated Attention; Interpretability; Feature Importance}

\maketitle

\section{Introduction}\label{sec:introduction}

\revise{Relational databases are the $de facto$ standard for storing and querying structured data that are critical to the operation of most businesses~\cite{scaleFM,relFD,fivm,factorizedla}.}
They capture a huge wealth of information that can be used for data-driven decision making, and for identifying risks and opportunities. Extracting insights from data for decision making requires advanced analytics.  In particular, deep learning, which is much more complex than statistical aggregation, has recently shown great promise.

Deep neural networks (DNNs) have led to breakthroughs in images, audio and text data~\cite{he2016deep, amodei2016deep, lai2015recurrent}.
DNNs such as CNNs~\cite{he2016deep} and LSTM~\cite{lai2015recurrent} are well-suited to particular data types for which they have been designed, e.g., CNNs for images and LSTM for sequential data.
\update{One major advantage of using DNNs is that their adoption obviates the need for manual feature engineering, which however may not produce meaningful results when applied to structured data as in relational tables.
Specifically, there are intrinsic correlations and dependencies among attribute values of structured data, and such feature interactions are essential for predictive analytics.
Although theoretically, a DNN can approximate any target function given sufficient data and capacity, the interactions captured for a conventional DNN layer are additive.
As a result, prohibitively large and increasingly inscrutable models that stack multiple layers with nonlinear activation functions in-between are required to model such multiplicative interactions~\cite{multiplicative,latentcross}.
Previous studies~\cite{afn,latentcross,polynomialnn,multiplicative} have also suggested that modeling such ``cross features'' implicitly with DNNs may take a substantial number of hidden units, which greatly increases the computational cost and at the same time renders it difficult to be interpreted.
}

\revise{Formally, structured data can be viewed as a logical table of $n$ rows (tuples/samples) and $m$ columns (attributes/features)~\cite{normalizedla,factorizedla}, which is extracted from relational databases via core relational operations such as select, project and join.
The predictive modeling is to learn the functional dependency (predictive function) of the \textit{dependent attribute} $y$ on the \textit{determinant attributes} $\mathbf{x}$, namely, $f: \mathbf{x} \rightarrow y$, where $\mathbf{x}$ is typically called the feature vector, and $y$ is the prediction target.}
\revise{The major challenge for the predictive modeling of structured data is actually how to model these dependencies and correlations between attributes, called {\em feature interactions}, via \textit{cross features}~\cite{scaleFM,afm,afn,latentcross,polynomialnn} that create new features by capturing interactions of raw input features.}
\revise{Specifically, a cross feature can be defined as $\prod_{i=1}^m x_i^{w_i}$, i.e, the product of input features with their respective \textit{interaction weights}.}
The weight $w_i$ specifies the contribution of the $i$-th feature to the cross feature; $w_i = 0$ deactivates the corresponding feature $x_i$ in the feature interaction, and the \textit{interaction order} of a cross feature refers to the number of its non-zero interaction weights.
\revise{Such cross features for relation modeling are central to the structured data learning, which enables the learning model to represent more sophisticated functions beyond simple linear aggregation of input features for predictive analytics.}

\update{A preferred alternative to DNNs in relational analytics would be modeling feature interactions explicitly, thereby obtaining generally better performance and interpretability in terms of feature attribution~\cite{xaisurvey,lime}.
}
However, the number of possible feature interactions is combinatorially large.
\revise{The central problem of explicit cross feature modeling is therefore how to identify the right set of features and meanwhile, specify the corresponding interaction weights.}
Most existing studies sidestep this problem by capturing cross features with the interaction order confined within a predefined maximum integer~\cite{fm,scaleFM,ffm,afm,factorizedla}.
However, the number of cross features still grows near exponentially as the maximum order increases.
AFN~\cite{afn} takes a step further by modeling cross features with logarithmic neurons~\cite{lnn}, each of which transforms features into a logarithmic space so that the powers of features are converted into learnable coefficients, specifically, $\exp(\sum_{i=1}^m w_i \log x_i)$.
\revise{In this manner, each logarithmic neuron can capture a specific feature interaction term of arbitrary orders.}
\revise{Nonetheless, AFN has inherent limitations that the input features of interaction terms are restricted to positive values on account of the logarithmic transformation, and the interaction order of each interaction term is unconstrained and remains static after training.}

\update{We contend that cross features should only account for certain input features, and feature interactions should be modeled dynamically in an instance-aware manner.
The rationale is that not all input features are constructive for the interaction term, and modeling with irrelevant features could simply introduce noise and thus reduce effectiveness and interpretability.
In particular, the deployment of learning models in real-world applications has highlighted the need for not only accuracy, but also efficiency and interpretability~\cite{lime,xaisurvey,biran2017explanation,miller2019explanation}.
Notably, understanding the general behavior and the whole logic of the learning model (\textit{global interpretability}) and providing reasons for making a specific decision (\textit{local interpretability})~\cite{xaisurvey,lime} is of paramount importance for high-stakes applications in critical decisions making, e.g., healthcare or finance~\cite{tracer}.
Despite their great predictive power, many ``black box'' models such as DNNs model inputs in an implicit way, which is inexplicable and sometimes can learn unintended patterns~\cite{mlsafety,adversarial}.
In this light, explicitly modeling feature relations with a minimum set of constituent features adaptively would yield desirable inductive biases for effectiveness, efficiency and interpretability.
}

\begin{figure}[t]
    \centering
    \includegraphics[width=0.84\linewidth]{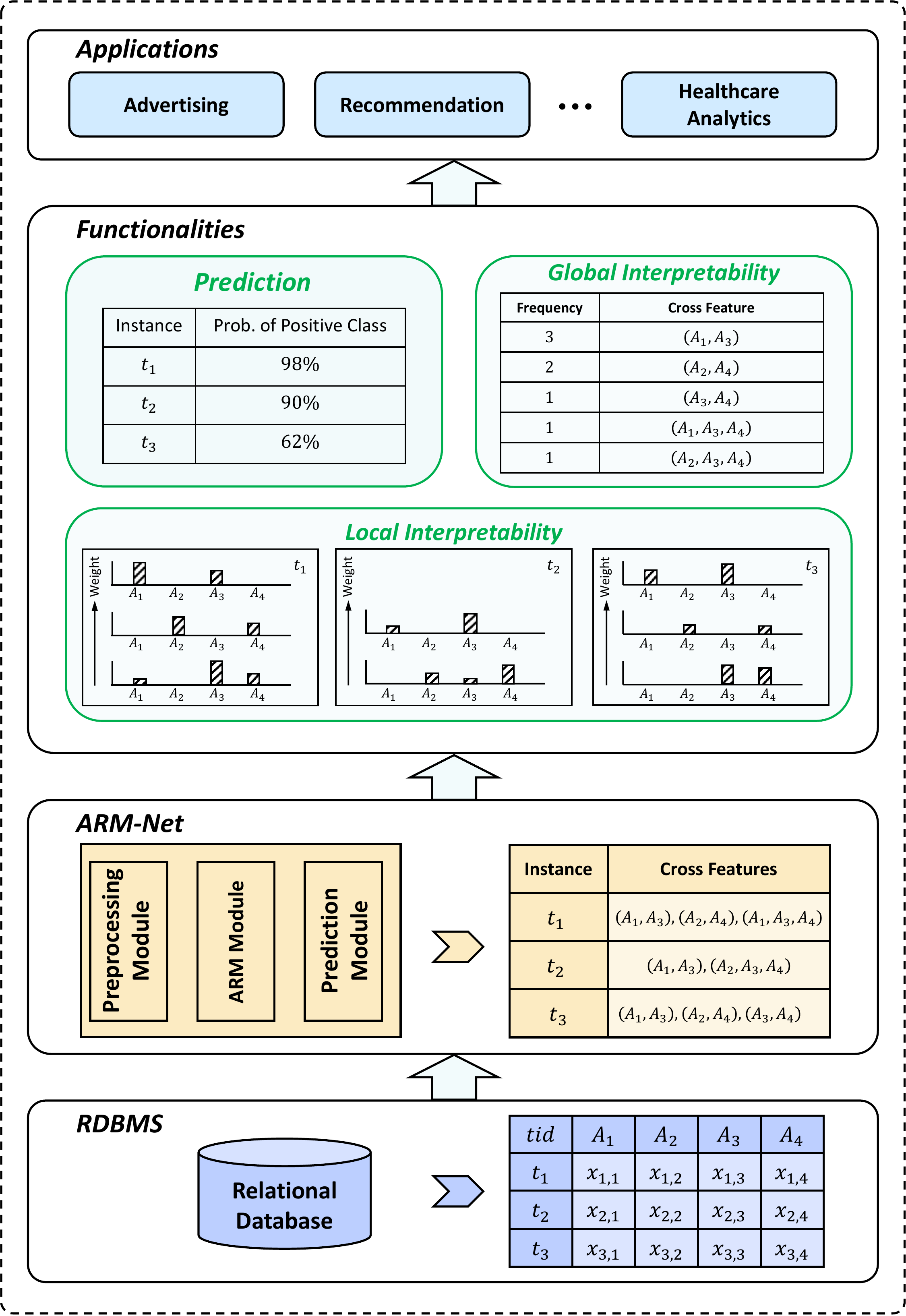}
    \caption{\revise{The overview of \framework based on the relation modeling of \name for structured data analytics.}}
    \label{fig:framework}
    \vspace{-6mm}
\end{figure}

In this paper, we present an \textbf{\underline{A}}daptive \textbf{\underline{R}}elation \textbf{\underline{M}}odeling \textbf{\underline{Net}}work (\name) for structured data, which models feature interactions of arbitrary orders selectively and dynamically.
To this end, we address the issue of adaptive feature selection for cross features with a gated attention mechanism, and model feature interaction weights and the interaction order dynamically with novel exponential neurons.
\revise{The key idea is to model feature interactions in the exponential space and determine the interaction weights dynamically based on the current input instance, i.e., each tuple of the structured data}.
In particular, exponential neurons transform input features into an exponential space, and then the interaction weights are dynamically determined by attention alignment followed by the gating with sparse softmax~\cite{entmax}.
As a consequence, each exponential neuron captures a specific cross feature of arbitrary orders with irrelevant features filtered by gated attention adaptively.
To the best of our knowledge, we are the first to propose an adaptive relation modeling network for structured data.
Based on \name, we develop a lightweight \textbf{\underline{A}}daptive \textbf{\underline{R}}elation \textbf{\underline{M}}odeling framew\textbf{\underline{OR}}k (\framework) for relational data analytics.
The overview of \framework is illustrated in Figure~\ref{fig:framework}.
In the training phase, \name is trained to model feature interactions in a selective and dynamic manner.
\update{
In the inference phase, given the input tuples, \framework supports key functionalities such as prediction, global interpretability and local interpretability for various structured data analytics.
Let us consider a use case when a company would like to make predictions about monthly sales, and a data table containing attribute fields of (\textit{month}, \textit{regionID}, \textit{storeID}, \textit{productID}) and some predictive targets \textit{total sales} are available.
In such an application, \framework can learn to predict the monthly sales target and 
disclose the cross features that have been used to make the prediction.
In this example, a particular store may perform better at selling a particular product locally, and all the stores may sell way more of a particular product in certain months/regions globally.
\framework is able to dynamically identify the interactions of these features and highlight such cross features in human understandable terms, on which the predictive analytics are based.
}
We summarize the main contributions as follows.

\begin{itemize}
    \item We propose exponential neurons and the gated attention mechanism for adaptive feature interaction modeling, which capture cross features selectively and dynamically.
    \item
    \revise{
    We develop an adaptive relation modeling framework, which during inference, takes structured data tuples as inputs and produces compact relational representations for predictive analytics, and meanwhile provides both global and local interpretation results for deriving insights. The implementation is available at GitHub\footnote{\href{https://github.com/nusdbsystem/ARM-Net}{https://github.com/nusdbsystem/ARM-Net}}.
    }
    \item We conduct extensive experiments on real-world datasets.
    The results confirm that our exponential neurons with gated attention can adaptively capture cross features of arbitrary orders, \revise{and our \framework consistently achieves superior prediction performance and better interpretability.}
\end{itemize}


In the remainder of this paper, we introduce preliminaries in Section~\ref{sec:preliminary}.
We present \name with a detailed introduction of its modules and optimization schemes in Section~\ref{sec:formulation}.
Experimental results are provided in Section~\ref{sec:experiment}.
We review related work in Section~\ref{sec:related work} and conclude the paper in Section~\ref{sec:conclusion}.

\section{Preliminaries}
\label{sec:preliminary}

In this section, we introduce the preliminaries of structured data, Logarithmic Neural Network (LNN) and sparse softmax.
\revise{We first discuss structured data considered in this paper.
Then, we present the two relevant techniques central to our \name, namely LNN for modeling higher-order interactions and sparse softmax for sparsely selecting informative features.}
Scalars, vectors and matrices are denoted as $x$, $\mathbf{x}$ and $\mathbf{X}$ respectively.

\vspace{1mm}
\noindent
\highlight{Structured Data.}
\revise{Most businesses to date rely on structured data for their data storage and predictive analytics~\cite{scaleFM,normalizedla,factorizedla,relFD}.
Relational Database Management System (RDBMS) has become the predominant database system adopted by the industry.}
Structured data (or relational data, tabular data)~\cite{normalizedla,factorizedla} refers to the type of data that can be represented in tables.
\revise{Structured data is generally stored in a set of tables (relations) $\{\mathbf{T}_1, \mathbf{T}_2, \dots\}$ of columns and rows, which can be extracted from a relational database with feature extraction queries, e.g., the projection, natural join, and aggregation of these tables in the database~\cite{normalizedla,factorizedla,embRel,relFD}}.
Each column conforms to a domain of certain constraints and corresponds to a specific feature in learning models.
Tables of structured data are linked to other tables via \textit{foreign key} attributes, i.e., the value of a column in a table relates to a unique row of another table.
For ease of discussion, we therefore formulate structured data as one logical table $\mathbf{T}$ of $n$ rows and $m$ columns.
\revise{Specifically, each row can be denoted as a tuple $(\mathbf{x}, y) = (x_1, x_2, \dots, x_m, y)$, where $y$ is the dependent attribute (prediction target), $\mathbf{x}$ is the determinant attributes (feature vector), and $x_i$ is the $i$-th attribute value which is either numerical or categorical.}
\revise{As existing solutions are not well-suited in terms of effectiveness, efficiency and interpretability, there has been a growing interest in designing learning models for structured data~\cite{scaleFM,gnnforrelnn,tabnet} and integrating predictive analytics into RDBMS~\cite{embRel,factorizedla,relFD,wwsigrec06}.
}

\vspace{1mm}
\noindent
\highlight{Logarithmic Neural Network.}
Feed-forward neural networks (FNNs) are known to be \textit{universal function approximators}.
Each neuron $y$ of FNN simply aggregates inputs $\mathbf{x}$ with corresponding learnable weights $\mathbf{w}$: $y = \sum_i^m w_i x_i$ followed by a non-linear activation.
Although FNNs can arbitrarily approximate any continuous function~\cite{approximation}, they are not well suited to model unbounded non-linear functions~\cite{lnn}, particularly, functions involving multiplication, division and power interactions between inputs.
Logarithmic Neural Networks~\cite{lnn,afn} (LNNs) instead approximate these higher-order interactions directly in the logarithmic space $y = \exp (\sum_i^m w_i \ln x_i) = \prod_i^m x_i^{w_i}$, where each logarithmic neuron $y$ operates on inputs $\mathbf{x}$ that are transformed into the logarithmic space.
As a consequence, multiplication, division and powers interactions between inputs can be converted into addition, subtraction and multiplication of the weights $\mathbf{w}$.
With such logarithmic transformation, the interaction weights between inputs can be determined adaptively, and each logarithmic neuron captures a specific interaction term, producing one cross feature accordingly.

\vspace{1mm}
\noindent
\highlight{Sparse Softmax.}\label{sec:sparsemax}
Softmax transformation~\cite{softmax} is a crucial function in neural network models, which maps an input vector $\mathbf{z}$ into a probability distribution $\mathbf{p}$ whose probabilities correspond proportionally to the exponentials of input values, i.e., $\mathrm{softmax}(\mathbf{z})_j = \frac{\exp (z_j)}{\sum_i \exp (z_i)}$.
The output of softmax can thus be subsequently used as the model output denoting the class probabilities or attention weights indicating the relative importance of inputs in the attention mechanism.
Denoting the $d$-dimension probability simplex as $\Delta^d \coloneqq \{ \mathbf{p} \in \mathbb{R}^d: \mathbf{p} \ge 0, \|\mathbf{p}\|_1 = 1 \}$, softmax can then be interpreted in the variational form with entropy~\cite{softmax_variational}: 

\begin{equation}\label{eq:softmax}
\mathrm{softmax}(\mathbf{z}) = \mathop{\mathrm{argmax}}_{\mathbf{p} \in \Delta^d} \{ \mathbf{p}^T \mathbf{z} + \mathbf{H}^{\mathbf{S}}(\mathbf{p}) \}
\end{equation}

\begin{figure*}[ht!]
    \centering
    \includegraphics[width=0.68\textwidth]{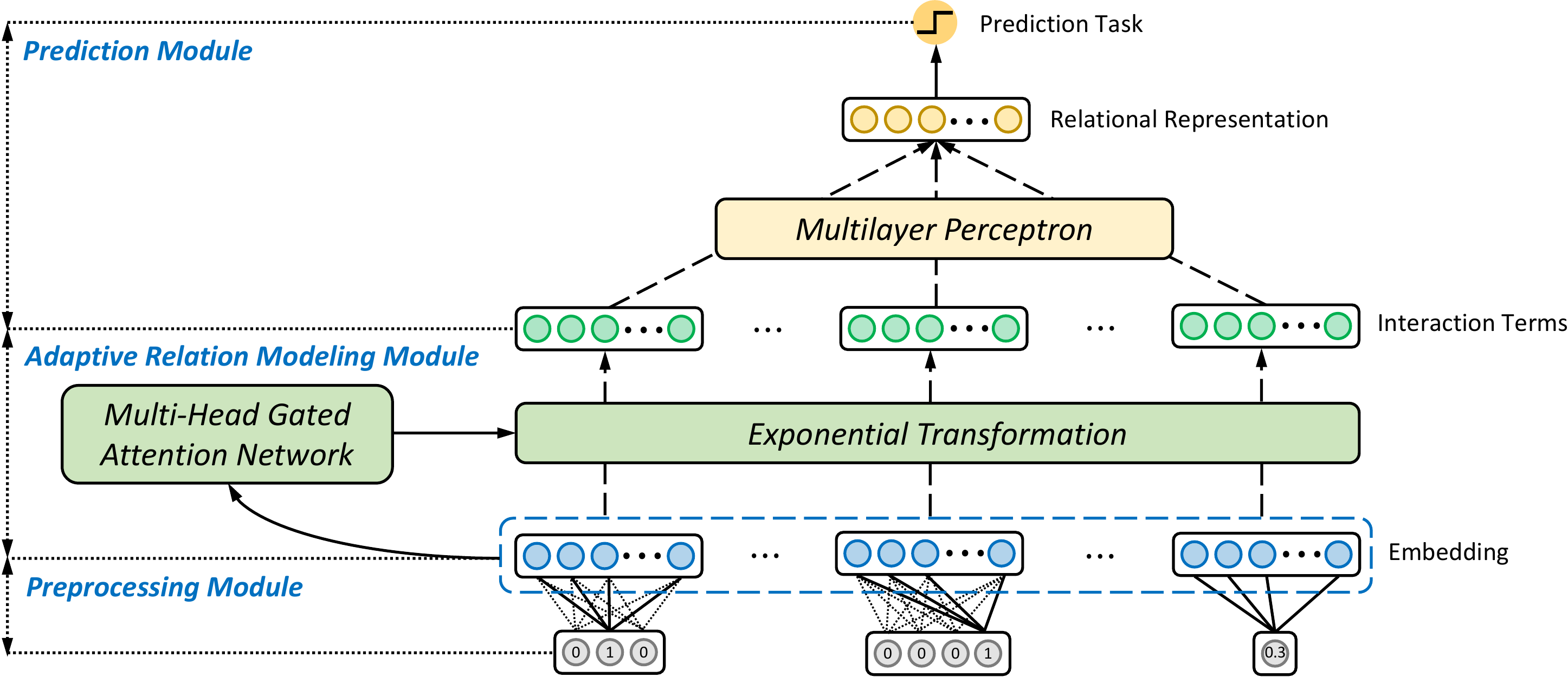}
    \caption{\revise{The overview of \name.}}
    \vspace{-3mm}
    \label{fig:armnet_overview}
    
\end{figure*}

\noindent
where $\mathbf{H}^{\mathbf{S}}(\mathbf{p}) = -\sum_j p_j \log p_j$ is the Shannon entropy.
The softmax function is straightforward to compute, differentiable and convex and is thus extensively used in neural networks.
However, softmax always assigns dense positive credits (probabilities) to all inputs, which compared with sparse credit assignment, is less interpretable and effective in attention alignment~\cite{entmax}.
To mitigate this issue, sparse softmax~\cite{sparsemax,entmax} is proposed to produce sparse distributions, by assigning zero probability to certain outputs.
Entmax~\cite{entmax} generalizes dense and sparse softmax with Tsallis $\alpha$-entropies $\mathbf{H}_{\alpha}^{\mathbf{T}}(\mathbf{p})$~\cite{tsallisentropy}:

\begin{equation}\label{eq:entmax}
\entmax(\mathbf{z}) = \mathop{\mathrm{argmax}}_{\mathbf{p} \in \Delta^d} \{ \mathbf{p}^T \mathbf{z} + \mathbf{H}_{\alpha}^{\mathbf{T}}(\mathbf{p}) \}
\end{equation}

\noindent
where $\mathbf{H}_{\alpha}^{\mathbf{T}}(\mathbf{p}) = \frac{1}{\alpha(\alpha-1)} \sum_j(p_j - p_j^{\alpha})$ if $\alpha \ne 1$, else $\mathbf{H}_{1}^{\mathbf{T}}(\mathbf{p}) = \mathbf{H}^{\mathbf{S}}(\mathbf{p})$.
With a larger $\alpha$, $\entmax$ tends to produce a sparser probability distribution and consequently, a larger proportion of input features will be filtered out.

\section{\name for Structured Data Analytics}
\label{sec:formulation}

\revise{In this section, we first present the overview of \name, which is the core component of \framework and is designed to adaptively model feature interactions for structured data.
We then elaborate on each module of \name and introduce the optimization scheme.
We further discuss the feature interactions captured in the relation modeling of \name for local and global interpretability, and analyze both effectiveness and efficiency.}

\subsection{Overview}\label{sec:overview}

The overview of \name is illustrated in Figure~\ref{fig:armnet_overview}.
The main intuition is that attribute value distributions have ``structure'', and feature interactions have ``structure''.
\update{This structure can be learned on a per-input basis for more effective and explicable interaction modeling.
We further contend that not all features are useful in interaction modeling, and capturing cross features by simply introducing more constituent features is neither efficient nor effective.}
Instead, we propose to capture cross features in a selective and dynamic manner.

Specifically, given input features $\mathbf{x}$, we propose to first transform each input feature into the exponential space.
Next, we adaptively model feature interaction terms with our proposed exponential neurons with a multi-head gated attention mechanism.
\update{Each exponential neuron is designed to model a specific cross feature of arbitrary orders explicitly, which is more effective and understandable by humans;
and the gated attention generates the interaction weights dynamically to filter noisy features selectively, which makes the modeling process more efficient, effective, and interpretable.
We then feed the relational representation, i.e., the captured cross features that model the interactions, to a final prediction module for the prediction task.
We call our model \name, since it performs Adaptive Relation Modeling.
}

\subsection{Architecture}\label{sec:architecture}

\subsubsection{Preprocessing Module}

The input of \name can be denoted as a vector $\mathbf{x}=[x_1, \dots, x_m]$ of $m$ attribute fields, which can be either categorical or numerical.
Each field of the raw input features $\mathbf{x}$ is then transformed into an embedding vector.
In particular, categorical fields are mapped into a low-dimensional latent space via an embedding lookup, i.e., $\mathbf{e}_i = \mathbf{E}_i[:, x_i], \mathbf{e}_i \in \mathbb{R}^{n_e}$, \revise{where $\mathbf{E}_i$ is the embedding matrix of the $i$-th attribute field, and $n_e$ is the embedding size.
Note that embedding vectors of $\mathbf{E}_i$ correspond to their respective categories in this field, and the number of different categories in a field can be extremely large for real-world applications.}
\revise{Meanwhile, numerical attribute fields also need to be transformed into embeddings of the same dimension: $\mathbf{e}_j = x_j \mathbf{\hat{e}}_j$,} where $x_j$ is a scalar feature value (e.g., scaled into $(0, 1]$), and $\mathbf{\hat{e}}_j$ is the embedding vector for the $j$-th feature.
\revise{In this way, we can obtain fixed size inputs, i.e., embedding vectors $\mathbf{E} = [\mathbf{e}_1, \mathbf{e}_2, \dots, \mathbf{e}_m]$, for the $m$ attribute fields uniformly as the model input.}

\subsubsection{\fullmname}
\begin{figure}
    \centering
    \includegraphics[width=0.72\linewidth]{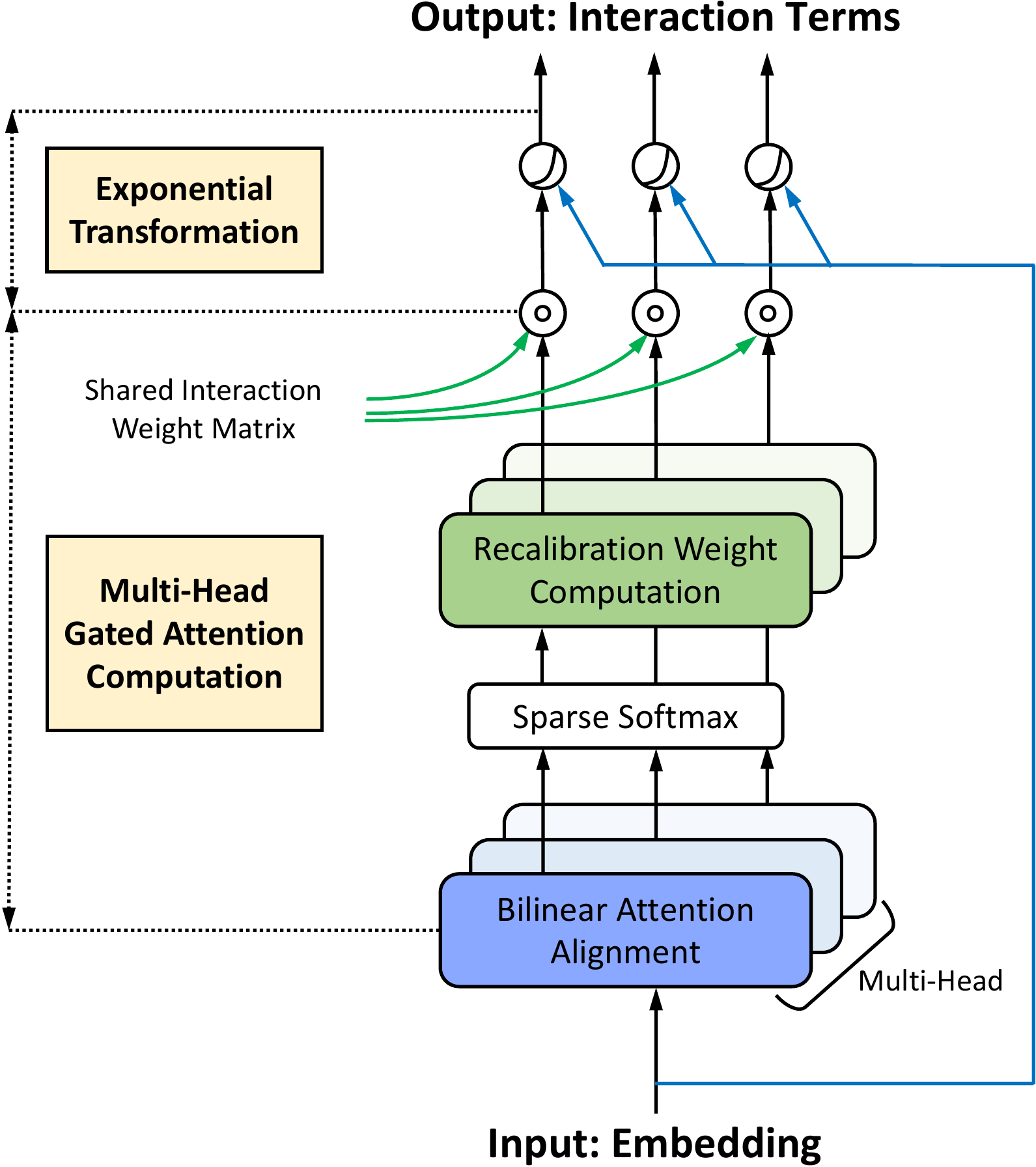}
    \caption{\fullmname of \name.}\vspace{-3mm}
    \label{fig:arm_module}
\end{figure}

To dynamically model the feature interactions, we propose \fullmname (\mname-Module) as illustrated in Figure~\ref{fig:arm_module}, in which we devise novel exponential neurons to model cross features of arbitrary orders.
Compared with LNNs~\cite{lnn,afn}, our exponential neurons relax the limitation of positive inputs for logarithmic neurons.
With exponential neurons, \mname-Module can determine the orders dynamically via a multi-head gated attention mechanism on a per-instance basis.
The detailed design of \mname-Module is introduced as follows.

\vspace{1mm}
\noindent
\powerpoint{Exponential Neurons.}
First, to address the limitation that inputs must be maintained positive for logarithmic neurons~\cite{lnn,afn}, we propose to process inputs in the exponential space instead of the logarithmic space, i.e., treat each input feature as the exponent of the natural exponential function.
We then propose the exponential neuron for exponential transformation accordingly:

\begin{equation}\label{eq:exponential_neuron}
\begin{split}
\mathbf{y}_i    &= \exp (\sum_{j=1}^m w_{ij} \mathbf{e}_j) \\
                &= \exp(\mathbf{e}_1)^{w_{i1}} \circ \exp(\mathbf{e}_2)^{w_{i2}} \circ \cdots \circ \exp(\mathbf{e}_m)^{w_{im}}
\end{split}
\end{equation}

\begin{equation}\label{eq:derivative}
\frac{\partial \mathbf{y}_i}{\partial \mathbf{e}_j} = \diag (w_{ij} \mathbf{y}_i), \frac{\partial \mathbf{y}_i}{\partial w_{ij}} = \mathbf{y}_i \circ \mathbf{e}_j
\end{equation}

\noindent
where $\circ$ denotes Hadamard product, $\exp(\cdot)$ and the corresponding exponent $w_{ij}$ are applied element-wise, and $\mathrm{diag}(\cdot)$ is the diagonal matrix function.
Feature interactions modeled in exponential neurons are thus based on feature embeddings transformed in the exponential space, namely, $\exp(\mathbf{e}_j)$, and the interaction weights are given by the weights $\mathbf{w}_i$ correspondingly.

\vspace{1mm}
\noindent
\powerpoint{Multi-Head Gated Attention.}
For the $i$-th exponential neuron $\mathbf{y}_i$, the power terms $\mathbf{w}_i = [w_{i1}, w_{i2}, \dots, w_{im}]$ are dynamically determined via a multi-head gated attention mechanism on a per-input basis.
\update{Such a selective attention mechanism guides each exponential neuron to attend to more informative features and suppress others for adaptive relation modeling, and due to its flexibility, the relation modeling process is more effective and parameter-efficient in capturing cross features than static approaches~\cite{fm,hofm,afn}.
}

\update{
To this end, we associate each exponential neuron with a learnable attention weight \textit{value vector} $\mathbf{v}_i \in \mathbb{R}^{m}$ shared across instances, which encodes the \textit{global attention weights} for each of the respective $m$ attribute field embeddings.}
Further, we propose to dynamically recalibrate attention values $\mathbf{v}_i$ of the $i$-th exponential neuron by the attention query alignment with embeddings of attribute fields followed by sparse softmax, so that noisy feature terms can be filtered, and the resultant interaction terms are only dedicated to informative features and thus more effective and interpretable.
Specifically, we associate each exponential neuron with another attention \textit{query vector} $\mathbf{q}_i \in \mathbb{R}^{n_e}$ to generate the attention recalibration weights dynamically via the bilinear attention alignment score~\cite{bilinearattention}:

\begin{equation}\label{eq:bilinear_attention}
\begin{split}
    \tilde{z}_{ij} &= \phi_{att} (\mathbf{q}_i, \mathbf{e}_j) = \mathbf{q}_i^{\mathsf{T}} \mathbf{W}_{att} \mathbf{e}_j \\
    \mathbf{z}_i &= \entmax(\mathbf{\tilde{z}}_i), \tilde{\mathbf{z}_i} \in \mathbb{R}^m
\end{split}
\end{equation}

\noindent
\update{where $\mathbf{W}_{att} \in \mathbb{R}^{n_e \times n_e}$ is the weight matrix for the bilinear attention function $\phi_{att}$ shared among exponential neurons}, and $\entmax$ is the sparse softmax introduced in Section~\ref{sec:sparsemax}, the sparsity of which increases with a larger $\alpha$.
The bilinear attention first projects the embedding vector $\mathbf{e}_j$ of the $j$-th attribute field dynamically into the query space of the $i$-th neuron, and then generates the attention score $\tilde{z}_{ij}$ by the alignment with the corresponding query vector $\mathbf{q}_i$.
We then calculate attention weights and obtain the dynamic feature interaction weights for each exponential neuron:

\begin{equation}\label{eq:attention_weight}
\begin{split}
    \mathbf{w}_i = \mathbf{z}_i \circ \mathbf{v}_i, \mathbf{w}_i \in \mathbb{R}^m
\end{split}
\end{equation}

\noindent
where the recalibration weights $\mathbf{z}_i$ is introduced as a \textit{gate} to deactivate noisy features and tone down less informative features for more effective interaction modeling and better interpretability, which is reminiscent of the output gate in LSTM~\cite{lstm} and the output channel recalibration in SENet~\cite{senet}.

Further, instead of modeling feature interactions with a single bilinear attention function, we adopt the multi-head attention mechanism~\cite{allyouneed,gat,reformer} with $K$ sets of different bilinear attention functions and attention value/query vectors accordingly, in order to support interaction modeling from different representation spaces.
Denoting the number of exponential neurons as $o$ for each attention head, we then have $o$ attention weight value vectors $\mathbf{V}^{(k)}=[\mathbf{v}_1^{(k)}, \dots, \mathbf{v}_{o}^{(k)}], \mathbf{V}^{(k)} \in \mathbb{R}^{m \times o}$ and query vectors $\mathbf{Q}^{(k)}=[\mathbf{q}_1^{(k)}, \dots, \mathbf{q}_{o}^{(k)}], \mathbf{Q}^{(k)} \in \mathbb{R}^{n_e \times o}$, \update{and obtain representations from the $o$ exponential neurons $\mathbf{Y}^{(k)} = [\mathbf{y}_1^{(k)}, \dots, \mathbf{y}_{o}^{(k)}]$ for the $k$-th attention function $\phi_{att}^{(k)}$.}
There are thus $K \cdot o$ feature interaction terms altogether, each of which models a specific feature interaction with the corresponding attention function and value/query vectors.

\subsubsection{Prediction Module}

With \mname-Module, we dynamically model vector-wise feature interactions via exponential neurons with the multi-head gated attention and obtain $K \cdot o$ cross features.
We vectorize and concatenate all the interaction terms $\{ \mathbf{Y}^{(1)}, \dots, \mathbf{Y}^{(K)} \}$ captured with exponential neurons to $\mathbf{y}$, where $\mathbf{y} \in \mathbb{R}^{K \cdot o \cdot n_e}$, then feed $\mathbf{y}$ to a multilayer perceptron (MLP) to further capture element-wise non-linear feature interactions, and obtain a vector $\mathbf{h}$ that encodes the relational representations:

\begin{equation}\label{eq:output_layer}
\mathbf{h} = \phi_{MLP}(\mathbf{y}),\ \phi_{MLP}: \mathbb{R}^{K \cdot o \cdot n_e} \mapsto \mathbb{R}^{n_h}
\end{equation}

\noindent
which is then forwarded to the final prediction layer:

\begin{equation}
\hat{\mathbf{y}} = \mathbf{W}_p \mathbf{h} + \mathbf{b}_p
\label{eq:pred_layer}
\end{equation}

\noindent
where $\mathbf{W}_p \in \mathbb{R}^{n_p \times n_h}$ and $\mathbf{b}_p \in \mathbb{R}^{n_p}$ are the weight and bias respectively, and $n_p$ corresponds to the number of the prediction targets of the learning task.
We propose to enhance the learning of \name with the ensemble of DNN,
which is introduced in the following section.

\subsection{Optimization}
\label{sec:optimization}

\noindent
\powerpoint{\name Training.}
\update{As shown in Equation~\ref{eq:output_layer} and Equation~\ref{eq:pred_layer}, \name can be adopted in various learning tasks, such as classification, regression with a proper objective function, e.g., MSE, cross entropy, etc.}
Take the binary classification tasks for example, the corresponding objective function is the binary cross entropy:

\begin{equation}\label{eq:logloss}
\logloss(\hat{\mathbf{y}}, \bar{\mathbf{y}}) = -\frac{1}{N} \sum_{i=1}^{N} \bar{y}_i \log \sigma(\hat{y}_i) + (1-\bar{y}_i) \log (1-\sigma(\hat{y}_i))
\end{equation}

\noindent
where $\hat{\mathbf{y}}$ and $\bar{\mathbf{y}}$ are the prediction labels and the ground truth labels respectively, \update{$N$ is the number of training instances, and $\sigma(\cdot)$ is the sigmoid function.}
With the objective function specified, \name can be trained effectively with popular gradient-based optimizers such as SGD, Adam~\cite{adam} and etc.

\vspace{1mm}
\noindent
\powerpoint{\name Ensemble with a DNN.}
\update{
Deep neural networks are known to be universal approximators with a sufficient number of hidden units and are powerful in capturing nonlinear feature interactions.
Following prior research~\cite{deepandcross,deepfm,xdeepfm,afn}, we further enhance the prediction output of \name with the deep interaction modeling of a DNN.
For the DNN model, we adopt another set of embedding vectors of attribute fields and vectorize them as the raw input to capture the fine-grained nonlinear feature interactions.}
In particular, the final prediction output of the ensemble of \name and a DNN is obtained by:

\begin{equation}\label{eq:final_output}
\hat{\mathbf{y}}_{+} = \mathbf{w}_1 \hat{\mathbf{y}}_{\name} + \mathbf{w}_2 \hat{\mathbf{y}}_{DNN} + \mathbf{b}_f
\end{equation}

\noindent
\update{where $\mathbf{w}_1$ and $\mathbf{w}_2$ are the ensemble weights for \name and DNN respectively, $\mathbf{b}_f \in \mathbb{R}^{n_p}$ is the bias, and $n_p$ is again the number of the prediction targets of the learning task.
The entire ensemble model can then be readily trained end-to-end by optimizing the objective function, e.g., Equation~\ref{eq:logloss}.}
We denote the ensemble model of \name and a DNN as \textbf{\nameplus}.

\subsection{Analysis and Discussion}
\label{sec:analysis_and_discussion}

\noindent
\highlight{Effectiveness.}
Most existing studies for feature interaction modeling either capture possible cross features statically with a predefined maximum interaction order~\cite{fm,ffm,afm}, or model cross features in an implicit manner~\cite{wideanddeep, shan2016deep}.
\update{However, different input instances should have distinct relations with varying constituent attributes.
Some relations are informative, whereas others might be simply noise.
Therefore, modeling cross features in a static manner is both parameter and computation inefficient, and less effective.
In particular, each exponential neuron output $y_i^{(k)}$ captures a specific cross feature of arbitrary orders and could potentially represent any combination of interacting features by deactivating other features.}
With the proposed exponential neurons and multi-head gated attention mechanism, \name can thus model feature interactions adaptively for better predictive performance.

\vspace{1mm}
\noindent
\highlight{\update{Interpretability.}}
Interpretability measures the extent to which decisions made by the model can be understood by human~\cite{xaisurvey,biran2017explanation, miller2019explanation}, which engenders user trust and provides new insights.
There exist general post-hoc interpretation approaches explaining how a \textit{black box} model works, including perturbation based~\cite{lime,shap}, gradient-based~\cite{attributionDNN} and attention-based~\cite{selvaraju2017grad} methods.
However, explanations produced by a different model is often not reliable, which can be misleading~\cite{xaisurvey,shap}.
\name instead follows \textit{transparent box design}~\cite{xaisurvey}, whose internal modeling process is more transparent and thus explainable for relational analytics.

Specifically, the interaction weights $\mathbf{w}_i^{(k)}$ for each feature interaction term $\mathbf{y}_i^{(k)}$ are derived from the attention values $\mathbf{v}_i^{(k)}$ shared across instances globally and are recalibrated by attention alignment dynamically for each instance.
Hence, the shared attention weight \textit{value vectors} encode the \textit{global interaction weights} before calibration for respective attribute fields over the instance population.
We can thus aggregate the absolute values of all the \textit{value vectors} of exponential neurons for \textit{global interpretability}, which indicates the general focus of \name on each attribute field in the data, namely the \textit{feature importance} of attribute fields.
Meanwhile, the proposed gated attention mechanism also encourages \textit{local interpretability}, which supports feature attribution on a per-input basis.
Note that each exponential neuron specifies a sparse set of attribute fields that are being used dynamically via the attention alignment.
Therefore, we can identify the cross features captured dynamically and similarly, obtain relative feature importance by aggregating the interaction weights of all exponential neurons for each instance.
The cross feature terms captured can also be analyzed globally/locally for understanding the internal modeling process.




\vspace{1mm}
\noindent
\highlight{Efficiency.}
Besides effectiveness and interpretability, model complexity is another important criterion for model deployment in real-world applications.
For simplicity of analysis and to reduce the number of hyperparameters, we set the embedding and attention vector sizes to $n_e$, and denote the parameter size of all MLPs in \name as $n_w$.
Recall that $m$, $K$, $o$ denotes the number of attribute fields, attention heads and exponential neurons for each attention head respectively.
\update{
Then for the preprocessing module, there are $O(Mn_e)$ feature embedding parameters with only $m$ attribute field embeddings being used for each instance, where $M$ is the number of distinct features, and $\frac{m}{M}$ is the overall sparsity.}
As $m$ is typically small and the preprocessing is simply embedding lookup and rescaling, the complexity of this part is negligible.

For the \mname-Module, the $K \cdot o$ exponential neurons of Equation~\ref{eq:exponential_neuron} can be computed in $O(K o m n_e)$;
the parameter size of value/query vectors is $O(K o n_e)$; and the computation complexity of the bilinear attention alignment for all the $m$ input embeddings is $O(K o m n_e^2)$ during training, which can be reduced to $O(K o m n_e)$ after training by precomputing the alignment of query vectors with respective attention weight matrices.
To further reduce complexity for certain applications, the shared attention weight matrix in Equation~\ref{eq:bilinear_attention} can be removed, resulting in single-head \name of complexity $O(K o m n_e)$ throughout.
For the prediction module, the complexity of the non-linear feature interaction function $\phi_{MLP}$ of Equation~\ref{eq:output_layer} is $O(n_w)$.
\update{Therefore, the overall parameter size and computational complexity for processing each input is $O(m n_e + n_w)$ and $O(K o m n_e + n_w)$ respectively during inference, which is linear to the number of attribute fields and thus is efficient and scalable.}


\section{Experiments}
\label{sec:experiment}

\begin{figure}
    \centering
    \includegraphics[width=0.68\linewidth]{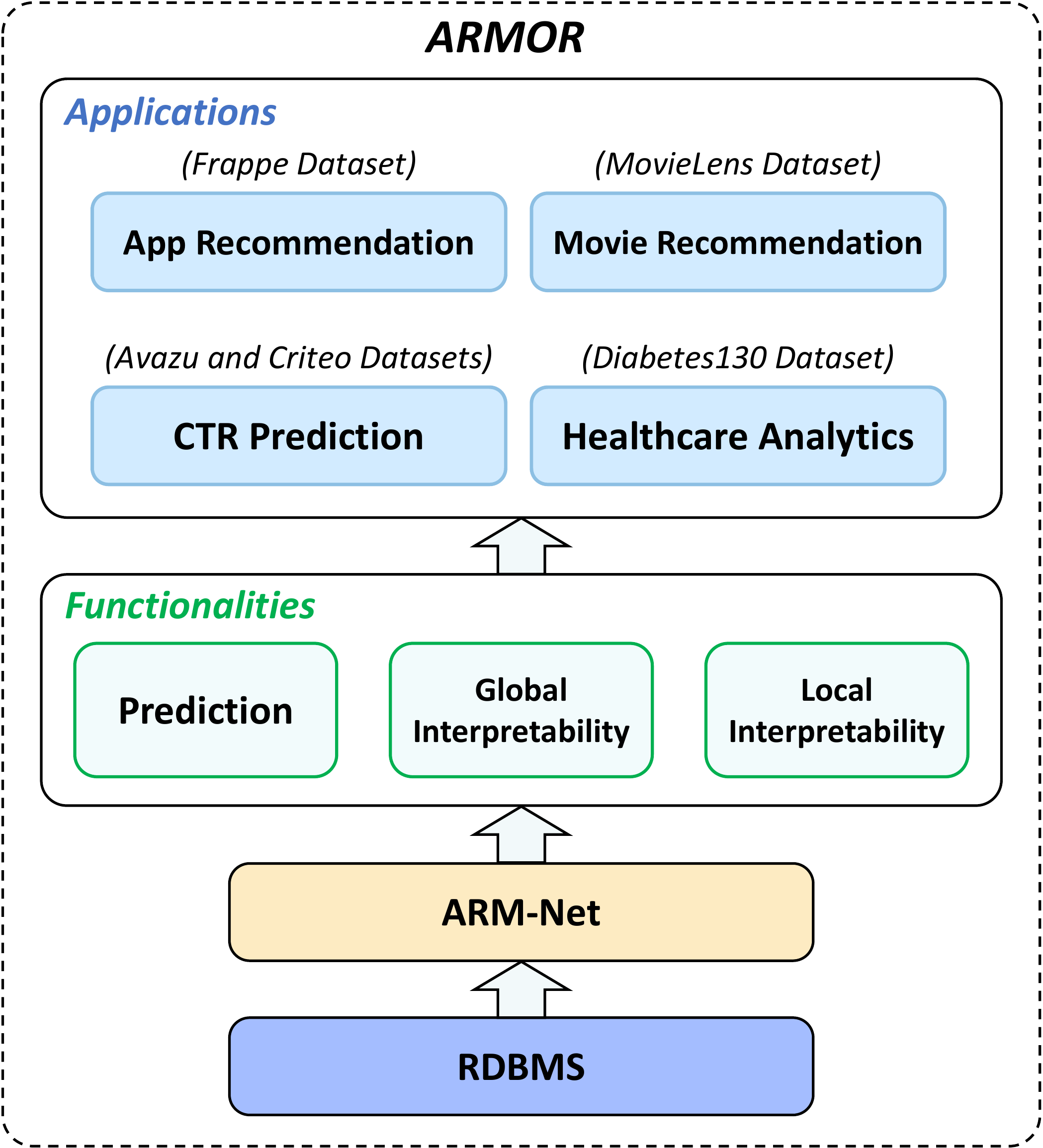}
    \caption{\revise{Predictive analytics using \framework.}}
    \label{fig:armor_exp}\vspace{-4mm}
\end{figure}

In this section, we evaluate the effectiveness, efficiency and interpretability of \framework.
\revise{Figure~\ref{fig:armor_exp} illustrates the overview of the experimental studies on \framework using five real-world datasets.}

\subsection{Experimental Setup}
\label{sec:exp_setup}

    
    

\subsubsection{Datasets and Applications}\label{sec:dataset}

\update{
We evaluate \framework with five real-world relational datasets on representative domains, namely app recommendation (\frappe\footnote{\href{https://www.baltrunas.info/research-menu/frappe}{https://www.baltrunas.info/research-menu/frappe}}), movie recommendation (\movielens\footnote{\href{https://grouplens.org/datasets/movielens/}{https://grouplens.org/datasets/movielens/}}), click-through rate prediction (\avazu\footnote{\href{https://www.kaggle.com/c/avazu-ctr-prediction}{https://www.kaggle.com/c/avazu-ctr-prediction}}, \criteo\footnote{\href{https://labs.criteo.com/2014/02/kaggle-display-advertising-challenge-dataset/}{https://labs.criteo.com/2014/02/kaggle-display-advertising-challenge-dataset/}}) and healthcare (\diabetes\footnote{\href{https://archive.ics.uci.edu/ml/datasets/diabetes+130-us+hospitals+for+years+1999-2008}{https://archive.ics.uci.edu/ml/datasets}}).
We summarize the statistics of the datasets in Table~\ref{tab:dataset_stats} with the hyperparameters searched for \name.
}

\vspace{1mm}
\noindent
\powerpoint{\frappe} is a real-world app recommendation dataset, which
contains a context-aware app usage log and consists of 96,203 tuples by 957 users for 4,082 apps used in different contexts.
\frappe samples two negative tuples for every one positive app usage log, leading to 288,609 tuples in total.
The learning task is to predict the app usage given the usage context, \revise{which includes 10 semantic attribute fields, such as the previous app usage count, weather, time, location, etc, with 5,382 different numerical/categorical embedding vectors.}

\vspace{1mm}
\noindent
\powerpoint{\movielens} contains the user tagging records of movies collected over various periods of time.
In this paper, we focus on personalized tag recommendation by preparing the dataset into the (user, movie, tag) format.
\revise{There are 2,006,859 tuples in total, and each tuple is a triplet of 3 categorical attribute fields of user ID, movie ID and tag, with 90,445 different categorical embedding vectors.}

\vspace{1mm}
\noindent
\powerpoint{\avazu} is a publicly accessible click-through rate (CTR) dataset provided by the mobile advertising platform Avazu.
In online advertising, CTR is an important metric for evaluating advertisement performance.
This dataset provides 11 days of data from Avazu for building models to predict whether a mobile ad will be clicked.
\revise{The dataset contains 40,428,967 tuples and 22 attribute fields with 1,544,250 different numerical/categorical embedding vectors such as app and device information.}
The dataset is split sequentially for training, validation and testing.

\vspace{1mm}
\noindent
\powerpoint{\criteo} is also a widely benchmarked CTR dataset, which contains attribute values and click feedback for millions of display ads.
Display advertising is of significant commercial value and one of the primary machine learning models deployed in the real world.
The learning task is to predict whether the user will click on a given ad under the page context.
\revise{The \criteo dataset comprises 45,840,617 tuples of 39 attribute fields with 2,086,936 different numerical/categorical embedding vectors, including 13 numerical attribute fields (mostly counts) and 26 categorical attribute fields.}

\vspace{1mm}
\noindent
\powerpoint{\diabetes} collects 10 years (1999-2008) clinical diabetes encounters at 130 US hospitals.
This dataset is presented for the analysis of a large clinical database
by examining the patterns of historical diabetes care, which might lead to improvements in providing safe and personalized healthcare for patients.
There are 101,766 encounters, each of which corresponds to a unique patient diagnosed with diabetes, and the learning task is the inpatient readmission prediction.
\revise{The dataset comprises 43 attribute fields with 369 different numerical/categorical embedding vectors, mainly including patient demographics and illness severity, such as gender, age, race, discharge disposition, primary diagnosis, etc.}

\begin{table}[t!]
    \small
    \centering
    \renewcommand{\arraystretch}{0.92}
    \caption{\update{Dataset statistics and best \name configurations.}
    }
    \label{tab:dataset_stats}
    \resizebox{0.96\columnwidth}{!}{
    \begin{tabular}{c|ccc|c}
    \toprule[1.8pt]
    
    Dataset     &   Tuples   &   Fields  &   Features       & \name Hyperparameters  \\ \midrule[1pt]
    \frappe     &   288,609     &   10      &   5,382       & $K=8, o=32, \alpha=2.0$   \\ \midrule[1pt]
    \movielens  &   2,006,859   &   3       &   90,445         & $K=1, o=16, \alpha=2.0$ \\ \midrule[1pt]
    \avazu      &   40,428,967  &   22      &   1,544,250      & $K=1, o=32, \alpha=1.5$ \\ \midrule[1pt]
    \criteo     &   45,302,405  &   39      &   2,086,936      & $K=4, o=64, \alpha=2.0$ \\ \midrule[1pt]
    \diabetes   &   101,766     &   43      &   369            & $K=1, o=32, \alpha=1.7$ \\
    
    \bottomrule[1.8pt]
    \end{tabular}
    }\vspace{-3mm}
\end{table} 

\subsubsection{Baseline Methods}\label{sec:baseline}
We compare \name with five categories of feature interaction modeling baselines:
(1) Linear Regression (LR) that linearly aggregates input attributes with their respective importance weights without considering feature interaction;
(2) models capturing second-order feature interactions, namely FM~\cite{fm}, AFM~\cite{afm};
\update{(3) models capturing higher-order feature interactions, namely HOFM~\cite{hofm}, DCN~\cite{deepandcross}, CIN~\cite{xdeepfm}, and AFN~\cite{afn};
(4) neural networks based methods, namely DNN, and graph neural networks GCN~\cite{gcn} and GAT~\cite{gat}.
(5) models that are an ensemble of explicit cross feature modeling and implicit feature interaction modeling by DNNs, namely Wide\&Deep~\cite{wideanddeep}, KPNN~\cite{pnn}, NFM~\cite{nfm}, DeepFM~\cite{deepfm}, DCN+~\cite{deepandcross}, xDeepFM~\cite{xdeepfm} and AFN+~\cite{afn}.}
We briefly introduce these baseline methods as follows.


\begin{itemize}[leftmargin=*]
\setlength\itemsep{0mm}

\item \powerpoint{LR} takes the raw field features as the input for prediction, which simply aggregates these features with respective weights.

\item \powerpoint{FM}~\cite{fm} explicitly models second-order feature interactions with the factorization technique for efficiency.

\item \powerpoint{AFM}~\cite{afm} enhances FM by capturing the relative importance of second-order cross features with attention dynamically.

\item \powerpoint{HOFM}~\cite{hofm} is a generalization of FM, which models higher-order feature interactions explicitly.

\item \powerpoint{DCN}~\cite{deepandcross} captures cross features by computing the feature cross of the input feature embeddings.

\item \powerpoint{CIN}~\cite{xdeepfm} models higher-order feature interactions by performing the compressed interaction with input embeddings iteratively.

\item \powerpoint{AFN}~\cite{afn} models feature interactions of arbitrary orders with logarithm neurons.

\update{
\item \powerpoint{DNN} captures fine-grained nonlinear feature interaction implicitly with a multilayer perceptron.

\item \powerpoint{GCN}~\cite{gcn} considers attribute fields as graph nodes and captures their interactions with neighboring nodes via graph convolution.

\item \powerpoint{GAT}~\cite{gat} also considers attribute fields as nodes and models their interaction with neighboring nodes with attention dynamically.
}

\item \powerpoint{Wide\&Deep}~\cite{wideanddeep} is an ensemble of LR and a DNN.

\item \powerpoint{KPNN}~\cite{pnn} captures feature interactions by computing kernel product between input embeddings and is enhanced with a DNN.

\item \powerpoint{NFM}~\cite{nfm} aggregates the element-wise product of all input pairs, and the bi-interaction features are enhanced with a DNN.

\item \powerpoint{DeepFM}~\cite{deepfm} is an ensemble model of FM and a DNN.

\item \powerpoint{DCN+}~\cite{deepandcross} is an ensemble model of DCN and a DNN.

\item \powerpoint{xDeepFM}~\cite{xdeepfm} is an ensemble model of CIN and a DNN.

\item \powerpoint{AFN+}~\cite{afn} is an ensemble model of AFN and a DNN.

\end{itemize}

\begin{table*}[ht]
    \small
    \centering
    \renewcommand{\arraystretch}{0.96}
    \caption{\update{Overall prediction performance with the same training settings.}}
    \label{tab:overall_resutls}
    
    \resizebox{1.98\columnwidth}{!}{
        \begin{tabular}{ c c c c c c c c c c c c}
    
    \toprule[1.5pt]
    \multirow{2}{*}{Model Class} & \multirow{2}{*}{Model} & \multicolumn{2}{c}{\frappe}  &   \multicolumn{2}{c}{\movielens}  &   \multicolumn{2}{c}{\avazu}  &   \multicolumn{2}{c}{\criteo}  &   \multicolumn{2}{c}{\diabetes}  \\
    &   &   AUC   &  Param  &   AUC   &   Param  &   AUC   &   Param  &   AUC   &   Param  &   AUC   &   Param    \\
    
    \midrule[0.5pt]
    First-Order &   LR  &   0.9336  &   5.4K    &  0.9215  &   90K   &  0.6900  &   1.5M  &  0.7741  & 2.1M   &   0.6701   &   370   \\
    
    \midrule[0.5pt]
    \multirow{2}{*}{Second-Order}    &   FM  &  0.9709  &   5.4K   &  0.9384  &   90K   &   0.6797  &   1.5M  &   0.7663  &   2.1M  &   0.6594   &   370 \\
    &   AFM  &  0.9665  &   5.7K   &  0.9473  &   91K   &   0.6857  &   1.6M  &   0.7847  &   2.1M  &   0.6774   &   7.6K \\
    
    \midrule[0.5pt]
    \multirow{5}{*}{Higher-Order}    &   HOFM  &   0.9778  &   170K  &  0.9435  &   2.9M   &   0.6919  &   18M  &   0.7788  &   107M  &   0.6714   &   11K \\
    &   DCN  &  0.9583  &   56K   &   0.9401  &   510  &   0.7460  &   3.3K  &   0.7959  &   6.6K  &   0.6765   &   2.2K \\
    &   CIN  &  0.9766  &   111K   &  0.9416  &   153K   &   0.6859  &   5.2M  &   0.7904  &   4.2M  &   0.6776   &   23K \\
    &   AFN  &  0.9779  &   3.1M   &   0.9470  &   242K  &   0.7456  &   3.3M  &   0.8061  &   7.8M  &   0.6778   &   306K \\
    &   \textbf{\name}  &  \textbf{0.9786}  &   867K   &   \textbf{0.9550}  &   140K  &   \textbf{0.7651}  &   147K  &   \textbf{0.8086}  &   1.5M  &   \textbf{0.6853}   &   102K \\
    
    \midrule[0.5pt]
    \multirow{3}{*}{NN-based}    &   DNN  &   \textbf{0.9787}  &   122K  &  \textbf{0.9540}  &   101K  &   0.7513  &   126K  &   \textbf{0.8082}  &   449K  &   0.6753   &   130K \\
    &   GCN  &  0.9732  &   1.6M   &   0.9404  &   365K  &   0.7506  &   964K  &   0.7984  &   2.2M  &   0.6828   &   4.0M \\
    &   GAT  &  0.9744  &   404K   &  0.9420  &   821K   &   \textbf{0.7525}  &   302K  &   0.8047  &   2.2M  &   \textbf{0.6846}   &   1.3M \\

    \midrule[0.5pt]
    \multirow{8}{*}{Ensemble}    &   Wide\&Deep  &  0.9762  &   127K   &  0.9477  &   192K   &   0.6893  &   1.7M  &   0.7913  &   2.5M  &   0.6626   &   130K \\
    &   KPNN  &  0.9787  &   140K   &  0.9546  &   102K   &   0.7514  &   195K  &   0.8089  &   893K  &   0.6794   &   582K \\
    &   NFM  &  0.9745  &   100K   &  0.9214  &   186K   &   0.6874  &   1.6M  &   0.7833  &   2.3M  &   0.6695   &   4.6K \\
    &   DeepFM  &  0.9773  &   127K   &  0.9481  &   192K   &   0.6891  &   1.7M  &   0.7899  &   2.5M  &   0.6683   &   131K \\
    &   DCN+  &  0.9786  &   213K   &  0.9553  &   192K   &   0.7487  &   168K  &   0.8079  &   703K  &   0.6844   &   227K \\
    &   xDeepFM  &  0.9775  &   538K   &  0.9481   &   215K  &   0.6913  &   1.8M  &   0.7917  &   5.0M  &   0.6659   &   55M \\
    &   AFN+  &  0.9790  &   365K   &  0.9563  &   343K   &   0.7524  &   3.0M  &   0.8074  &   8.0M  &   0.6825   &   741K \\
    &   \textbf{\nameplus}  &   \textbf{\underline{0.9800}}  &   263K  &   \textbf{\underline{0.9592}}  &   217K  &   \textbf{\underline{0.7656}}  &   339K  &   \textbf{\underline{0.8090}}  &   1.3M  &   \textbf{\underline{0.6871}}   &   1.7M \\
    
    \bottomrule[1.5pt]
        \end{tabular}
    }\vspace{-3mm}
\end{table*}

\subsubsection{Evaluation Metrics}

We evaluate the effectiveness of \framework with the metrics of AUC (the area under the ROC curve, higher is better), and $\logloss$ (cross entropy, lower is better).
For both AUC and $\logloss$, an improvement at 0.001 level is considered to be significant on the adopted benchmark datasets~\cite{deepfm,afn}.
\update{We split the dataset in 8:1:1 for training, validation and testing respectively, and report the mean values of the evaluation metrics from five independent runs with early stopping on the validation set.}

\subsubsection{\update{Hyperparameter Settings}}\label{sec:hyper_settings}

For a fair comparison, we evaluate the effectiveness of all the models with the same training settings by fixing the embedding size to 10 consistently across datasets and sharing the best searched configurations of DNN for all ensemble models.
Particularly, we obtain the best DNN depth from $1 \unsim 8$ and width from $100 \unsim 800$.
The number of GCN/GAT layers are best of $\{1, 2, \dots, 8\}$; the interaction orders for HOFM, DCN, CIN, DCN+ and xDeepFM are searched in $1 \unsim 8$, and the number of features/neurons for GCN, GAT, CIN, xDeepFM and AFN are searched in $\{10, 25, 50, 100, \dots, 1600\}$.

For \name in particular, the sparsity $\alpha$ is searched in $1.0 \unsim 3.0$; and the number of attention heads $K$ and exponential neurons per head $o$ are grid searched with the constraint $ K\cdot o \le 1024$.
We perform sensitivity analysis on these key hyperparameters in Section~\ref{sec:exp_ablation} and report the results of \name and the best searched baseline models in Table~\ref{tab:overall_resutls}.

\subsubsection{Implementation Details}
\hfill

\vspace{1mm}
\noindent
\powerpoint{Training Details.}
\update{
We adopt Adam~\cite{adam} optimizer with a learning rate searched in $0.1 \unsim 1e$-3 and generally a batch size of 4096 for all the models.}
In particular, we adopt a batch size of 1024 for the smaller dataset \diabetes, and evaluate the larger dataset \avazu every 1000 training steps.

\vspace{1mm}
\noindent
\powerpoint{Experimental Environment.}
\update{The experiments are conducted in a server with Xeon(R) Silver 4114 CPU @ 2.2GHz (10 cores), 256G memory and GeForce RTX 2080 Ti.
Models are implemented in PyTorch 1.6.0 with CUDA 10.2.}

\subsection{\update{Effectiveness}}
\label{sec:exp_effectiveness}

\vspace{1mm}
\noindent
\powerpoint{\update{Explicit Interaction Modeling with Single Models.}}
We first compare \name with single baseline models that capture first-order, second-order and higher-order cross features explicitly.
In Table~\ref{tab:overall_resutls}, we summarize the overall experimental results measured in AUC with respective parameter sizes used during inference.
Based on these results, we have the following findings.

\begin{figure}[t]
    \centering
    \includegraphics[width=\linewidth]{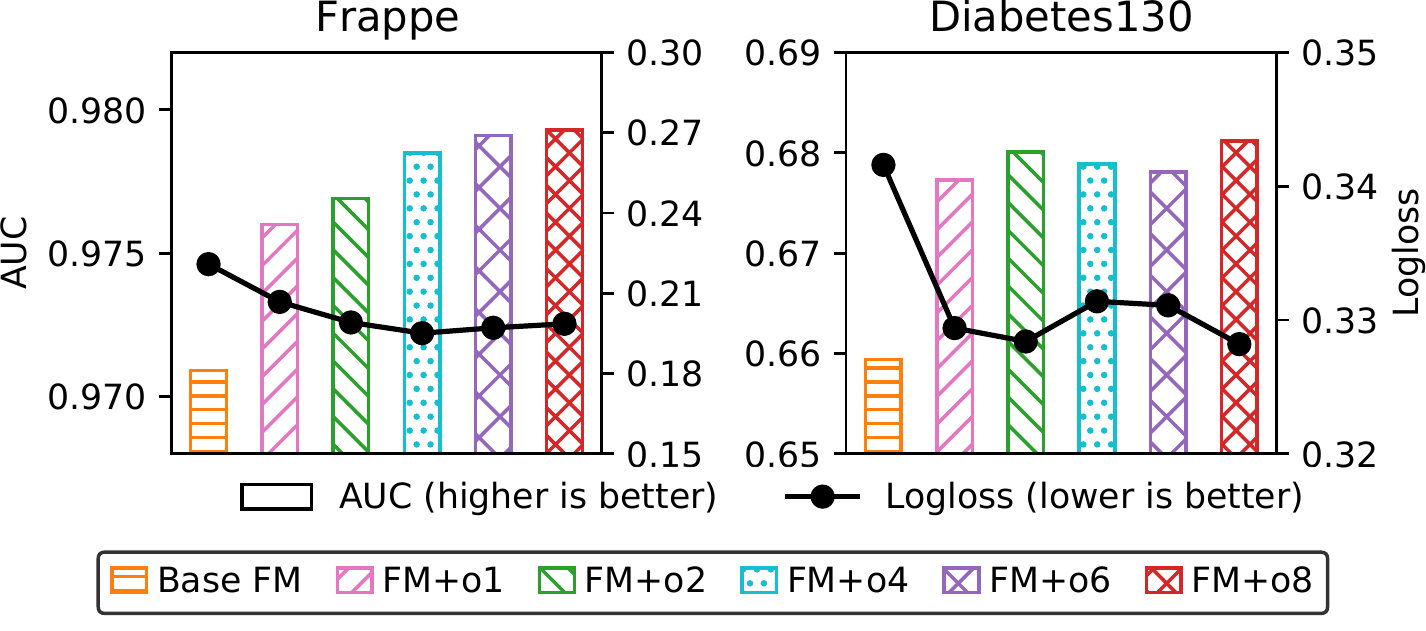}
    \caption{\update{The effectiveness of cross features captured by \name in enhancing FM (0 to 8 features respectively).
    }
    }
    \label{fig:synthetic}\vspace{-6mm}
\end{figure}

First, \name consistently outperforms explicit interaction modeling baselines in terms of AUC.
The better predictive performance confirms the effectiveness of \name across datasets and domains, including app recommendation (\frappe), movie tagging (\movielens), click-through rate prediction (\avazu and \criteo) and medical readmission (\diabetes).
Second, we find that higher-order models, e.g., HOFM and CIN, generally achieve better prediction performance than lower-order ones such as LR and FM, which validates that higher-order cross features are indeed essential for prediction, and the absence of which greatly reduces the modeling capacity.
Third, both AFN~\cite{afn} and \name significantly outperform baseline models of a fixed order, which verifies the efficacy of modeling feature interaction of arbitrary orders in an adaptive and data-driven manner.
Lastly, \name obtains noticeably higher AUC than the generally best-performing baseline model AFN.

\begin{table}[t!]
    \small
    \centering
    \renewcommand{\arraystretch}{1.0}
    \caption{\update{Training and inference efficiency of \name.}}
    \label{tab:efficiency}
    
    \resizebox{1.0\columnwidth}{!}{
        \begin{tabular}{ c c c c c c c c}
    
    \toprule[1.5pt]
    \multirow{2}{*}{Dataset} & \revise{Attribute} & \multicolumn{2}{c}{Training Throughput}  &   \multicolumn{2}{c}{Inference Throughput} & Avg GPU \\
    &  \revise{Field}  &   CPU   &   GPU  &   CPU   &   GPU   &  Speedup   \\
    
    \midrule[0.5pt]
    \movielens &   3   &   5,454  &   131,864    &  8,005  &   189,460  &  23.92x \\
    \frappe &   10   &   2,102  &   72,612    &  2,718  &   91,572 &   34.12x \\
    \avazu &   22     &   1,065  &   41,313    &  1,284  &   48,055 &   38.11x \\
    \criteo &   39    &   661  &   24,717    &  781  &   26,543  &  34.93x  \\
    \diabetes &   43     &   581  &   23,547    &  746  &   23,615 &  36.10x  \\
    \bottomrule[1.5pt]
        \end{tabular}
    }\vspace{-6mm}
\end{table}

The better performance of \name can be mainly ascribed to the exponential neurons and the gated attention mechanism.
Specifically, the restriction of positive inputs for logarithmic transformation in AFN limits its representation, while \name circumvents this problem by modeling feature interactions in the exponential space instead.
Further, instead of modeling interactions statically as in AFN, the multi-head gated attention of \name selectively filter noisy features and generates the interaction weights accounting for the characteristics of each input instance dynamically.
Therefore, \name can capture more effective cross features for better prediction performance on a per-input basis
Due to such runtime flexibility, \name is also more parameter efficient.
As shown in Table~\ref{tab:dataset_stats}, the best \name takes only dozens to a few hundreds of exponential neurons for different scale datasets, while the best AFN generally takes over a thousand neurons to obtain its best results, e.g., 32 as compared with 1600 for \name and AFN respectively on the large dataset \avazu.

\vspace{1mm}
\noindent
\powerpoint{\update{NN-based Models and Ensemble Models.}}
The results of NN-based models and DNN ensemble models are shown in Table~\ref{tab:overall_resutls}.
From these results, we can summarize the following key findings.
(1) Despite not explicitly modeling feature interactions, the best NN-based models generally achieve strong prediction performance against other single model baselines.
Particularly, the attention-based graph network GAT obtains noticeably higher AUC than other single models on \avazu and \diabetes.
However, their performance is not as consistent as \name, which varies greatly across different datasets, e.g., GAT performs much worse than DNN and \name on \frappe and \movielens.
(2) Model ensembles with a DNN significantly improve their respective predictive performance.
This can be observed consistently throughout the baseline models, e.g., DCN+, xDeepFM and AFN+, which suggests that the nonlinear interactions captured by DNNs are complementary to explicitly captured interactions.
(3) \name achieves comparable performance with DNN, and \nameplus further improve the performance considerably, which obtains the overall best performance across all the benchmark datasets.
In a nutshell, these results further confirm the effectiveness of \name in modeling feature interactions of arbitrary orders in a selective and dynamic manner.

\subsection{Efficiency and Ablation Studies}
\label{sec:exp_ablation}
\hfill

\begin{figure}
    \centering
    \includegraphics[width=1.\linewidth]{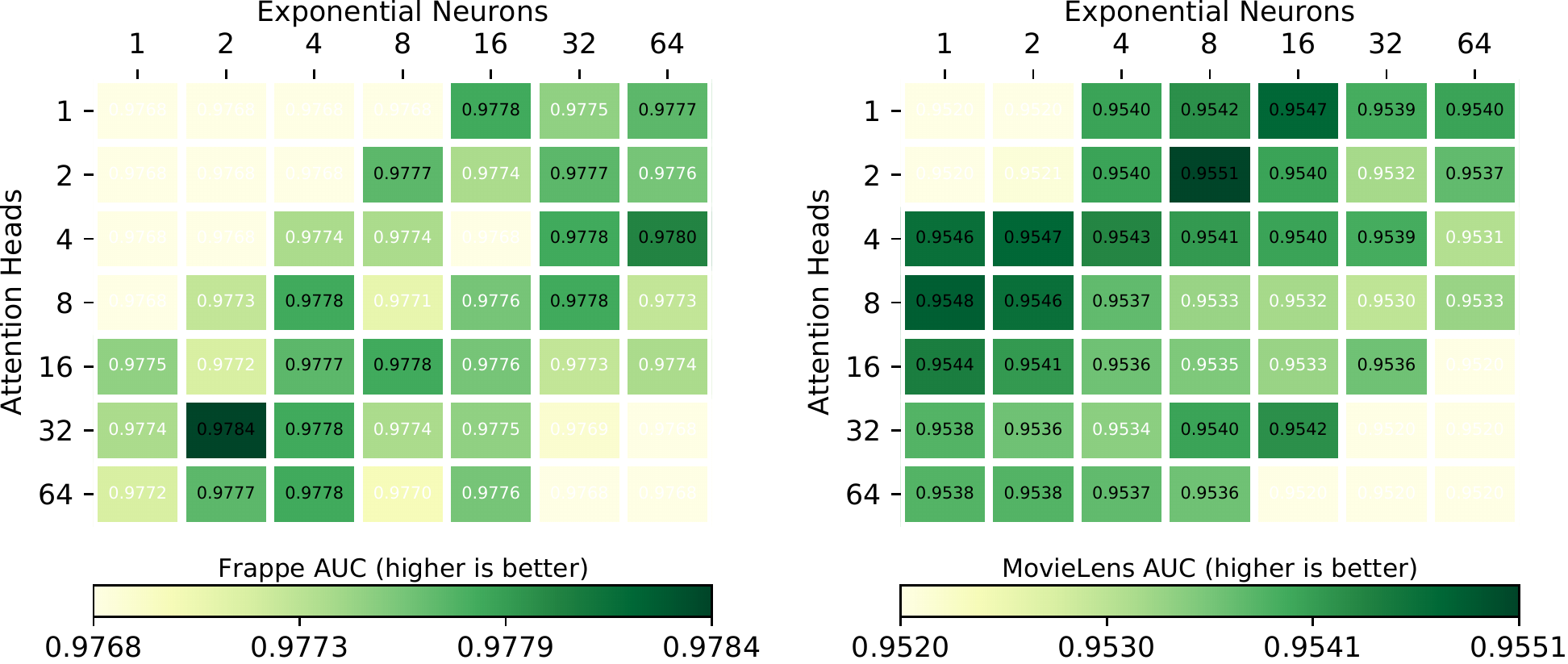}
    \caption{\update{Sensitivity analysis of the number of attention heads and exponential neurons per head ($\alpha=1.7$).}}
    \label{fig:multi_head_attn}\vspace{-3mm}
\end{figure}

\begin{figure}
\centering
\begin{subfigure}{.25\textwidth}
  \centering
  \includegraphics[width=.98\linewidth]{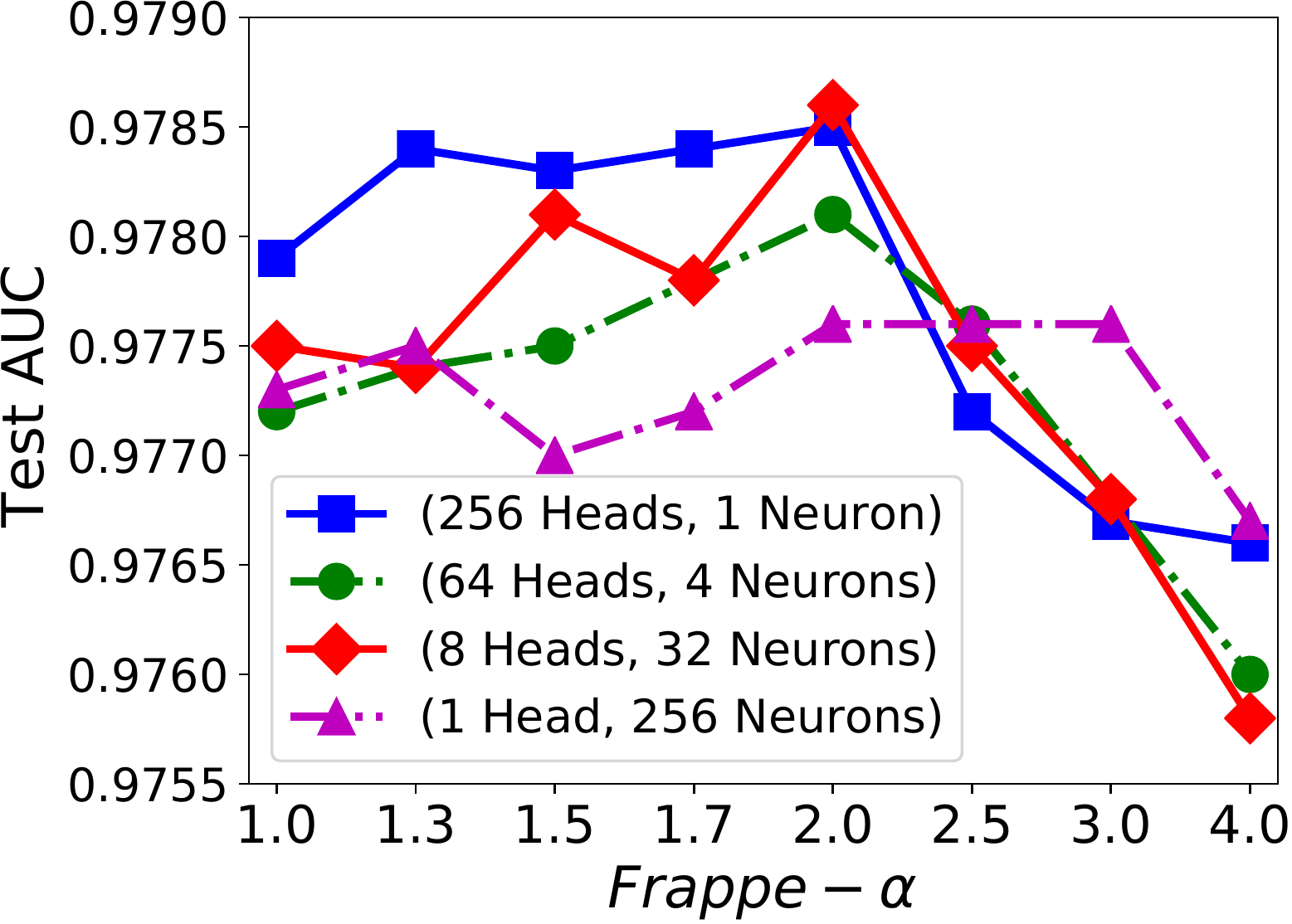}
\end{subfigure}%
\begin{subfigure}{.24\textwidth}
  \centering
  \includegraphics[width=0.96\linewidth]{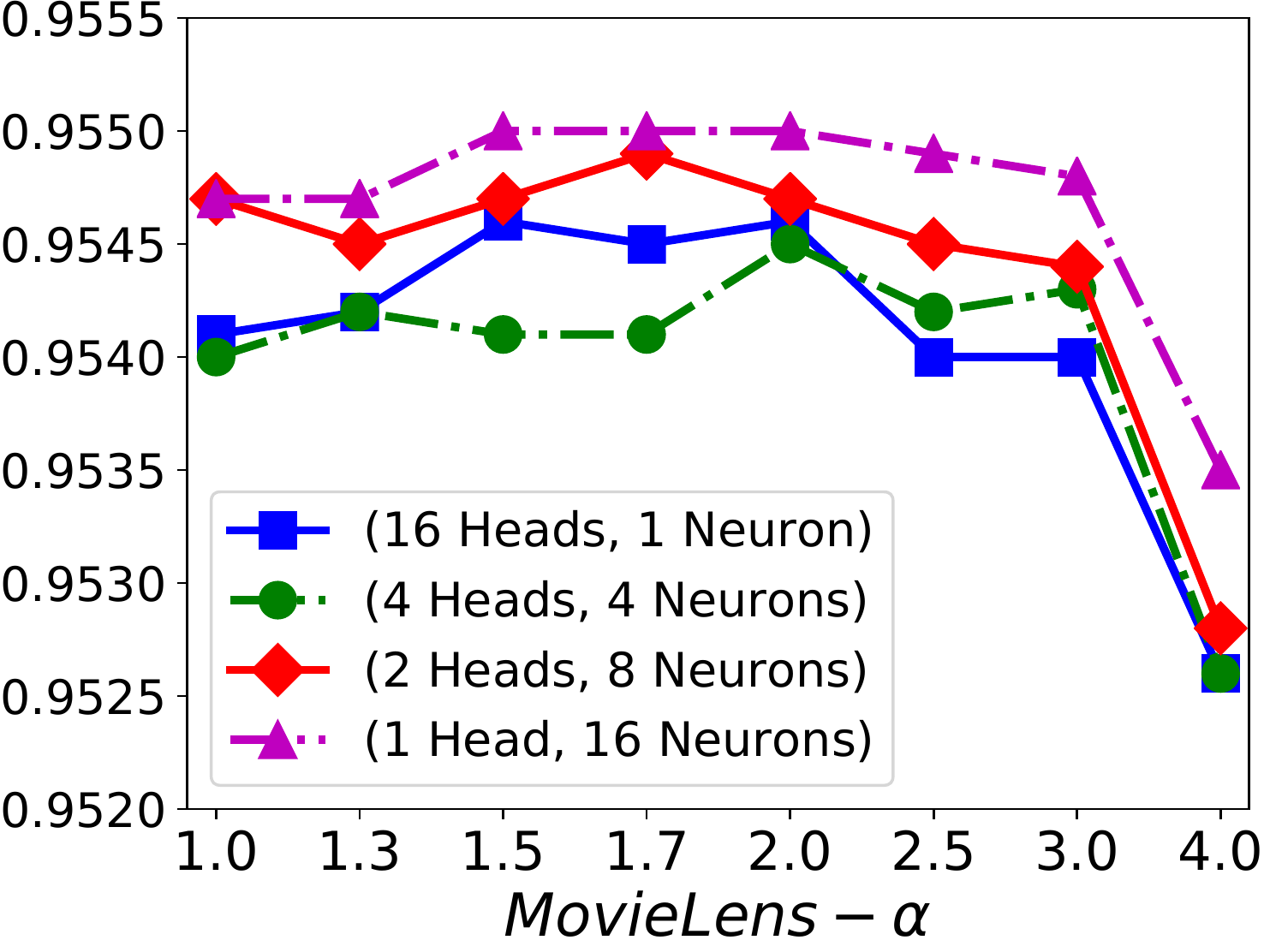}
\end{subfigure}
\caption{\update{The impact of varying the sparsity $\alpha$ on the prediction performance of different $K \cdot o$ configurations.}}
\label{fig:alpha_sensitivity}\vspace{-6mm}
\end{figure}

\vspace{1mm}
\noindent
\powerpoint{\update{Efficiency.}}
We further evaluate the training and inference efficiency of \name on the adopted benchmark datasets of different attribute fields size ($m$) in Table~\ref{tab:efficiency}, which shows the \textit{training/inference throughput} (the number of training/inference tuples per second).
We consistently adopt $K=4$, $o=64$ and $n_e=10$ for the benchmark \name, and train the model on one CPU or GPU.
The mini-batch size is set to 16,384 consistently for all datasets to saturate the computation of the GPU.

From Table~\ref{tab:efficiency}, we can observe that \name is rather efficient in both training and inference, whose throughputs are high and decrease linearly with the size of attribute fields $m$.
This is in line with our analysis in Section~\ref{sec:analysis_and_discussion} that the computational complexity of \name scales linearly to $m$.
Further, we find that GPU can considerably speed up both training and inference, with a ratio from 23.92x to 38.11x for the benchmark \name.
The speedup suggests that the efficiency of \name can be further improved by adopting more powerful hardware or more computational resources.

\vspace{1mm}
\noindent
\powerpoint{\update{Enhancing FM with Exponential Neurons.}}
To further evaluate the effectiveness of the cross features captured by the exponential neurons, we collect the feature interaction terms in \name and augment the learning of the factorization machine model (FM~\cite{fm}) with these interaction features on top of the FM feature embeddings.
Figure~\ref{fig:synthetic} shows the results measured in AUC and $\logloss$ of the baseline FM model and FMs enhanced with different numbers of exponential neurons on \frappe and \diabetes.

We can observe that the interaction features dynamically captured with the exponential neurons consistently improve the prediction performance of FM by a large margin, which indicates that the cross features captured by \name are orthogonal to the FM features, and the integration leads to better interaction modeling.
Remarkably, the improvement is significant even with only one exponential neuron, which enhances the AUC from 0.9709 of the Base FM to 0.9760 of FM+o1.
Further, the prediction performance increases as we integrate more cross features encoded by exponential neurons.
These findings also corroborate the modeling efficiency with the gated attention mechanism, which greatly reduces redundancy
thanks to the runtime interaction order determination.

\begin{table}[t]
    \small
    \centering
    \caption{Top Global Interaction Terms for \frappe.}
    \label{tab:frappe_interaction_terms}
    \begin{tabular}{cc|c}
    \toprule[1.8pt]
    
    Frequency     &   Orders   &   Interaction Term \\ \midrule[1pt]
    3.71     &   3     &   (weekday, location, is\_free)      \\
    3.52     &   3     &   (user\_id, item\_id, is\_free)      \\
    3.37     &   3     &   (item\_id, weekend, is\_free)      \\
    3.36     &   3     &   (item\_id, is\_free, city)      \\
    3.32     &   3     &   (user\_id, weekend, is\_free)      \\
    3.24     &   3     &   (user\_id, is\_free, city)      \\
    3.22     &   2     &   (item\_id, is\_free)      \\
    3.00     &   3     &   (user\_id, item\_id, weather)      \\
    \bottomrule[1.8pt]
    \end{tabular}\vspace{-2mm}
\end{table} 

\begin{figure}
    \centering
    \includegraphics[width=1.\linewidth]{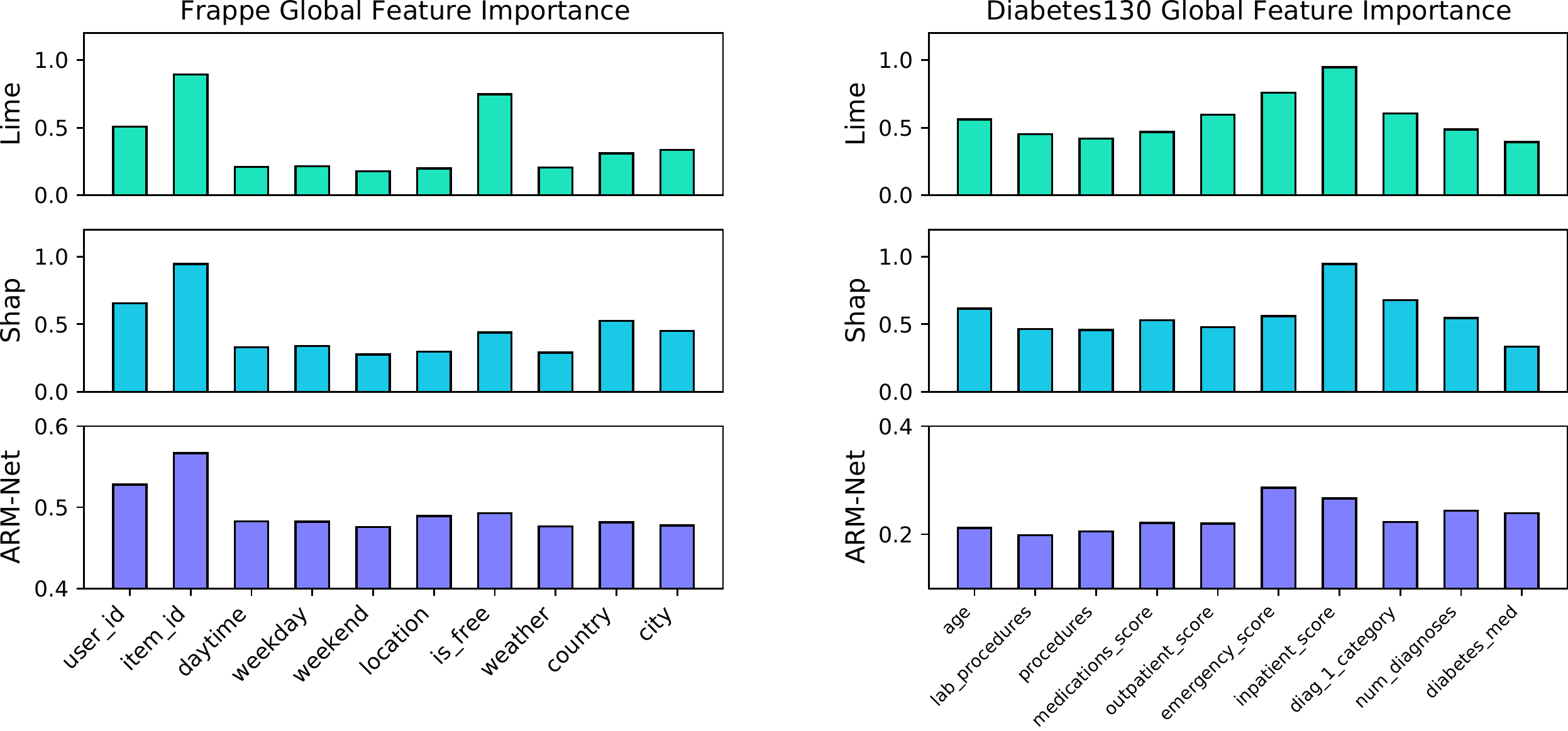}\vspace{-3mm}
    \caption{\update{Global feature attribution with Lime, Shape and \name respectively on \frappe and \diabetes.}}\vspace{-6mm}
    \label{fig:global_frappe_diabetes}
\end{figure}

\vspace{1mm}
\noindent
\powerpoint{\update{Multi-head Attention and Sparsity.}}
As discussed in Section~\ref{sec:analysis_and_discussion} and Section~\ref{sec:hyper_settings}, $K$, $o$, and $\alpha$ are the key hyperparameters of \name.
Specifically, $K$ and $o$ are the numbers of attention heads and exponential neurons per head respectively, and $\alpha$ controls the sparsity of $\alpha$-entmax for the sparse feature selection in the multi-head attention alignment.
We then evaluate these hyperparameters on \frappe and \diabetes and show the sensitivity analysis of $K$ and $o$ in Figure~\ref{fig:multi_head_attn} and the impact of sparsity $\alpha$ on different configurations of $K \cdot o$ in Figure~\ref{fig:alpha_sensitivity}.

In Figure~\ref{fig:multi_head_attn}, we can notice that \name obtains relatively stable and high prediction performance under different configurations of $K \cdot o$.
In particular, AUC scores achieved by \name on \movielens are consistently higher than 0.9470 by the best-performing baseline single model AFN~\cite{afn}.
Further, we can also find that simply increasing $K$ or $o$ may not necessarily lead to higher prediction performance.
This finding suggests that modeling from a moderate number of representation spaces with the multi-head attention mechanism is beneficial, and sharing the linear attention alignment weight within each attention head also facilitates learning and meanwhile, greatly reduces the parameter sizes.

Based on the results of varying sparsity on different $K \cdot o$ configurations in Figure~\ref{fig:alpha_sensitivity}, we find that a moderate sparsity $\alpha$ consistently leads to better prediction performance.
Specifically, with a sparsity of 2.0 and 1.7 for \frappe and \movielens respectively, \name achieve noticeably higher AUC than the corresponding architectures with dense softmax, namely $\alpha=1.0$, across different configurations.
These results support our contention that the proposed sparse attention mechanism can help filter noisy features dynamically for more effective interaction modeling.

\begin{figure}
    \centering
    \includegraphics[width=\linewidth]{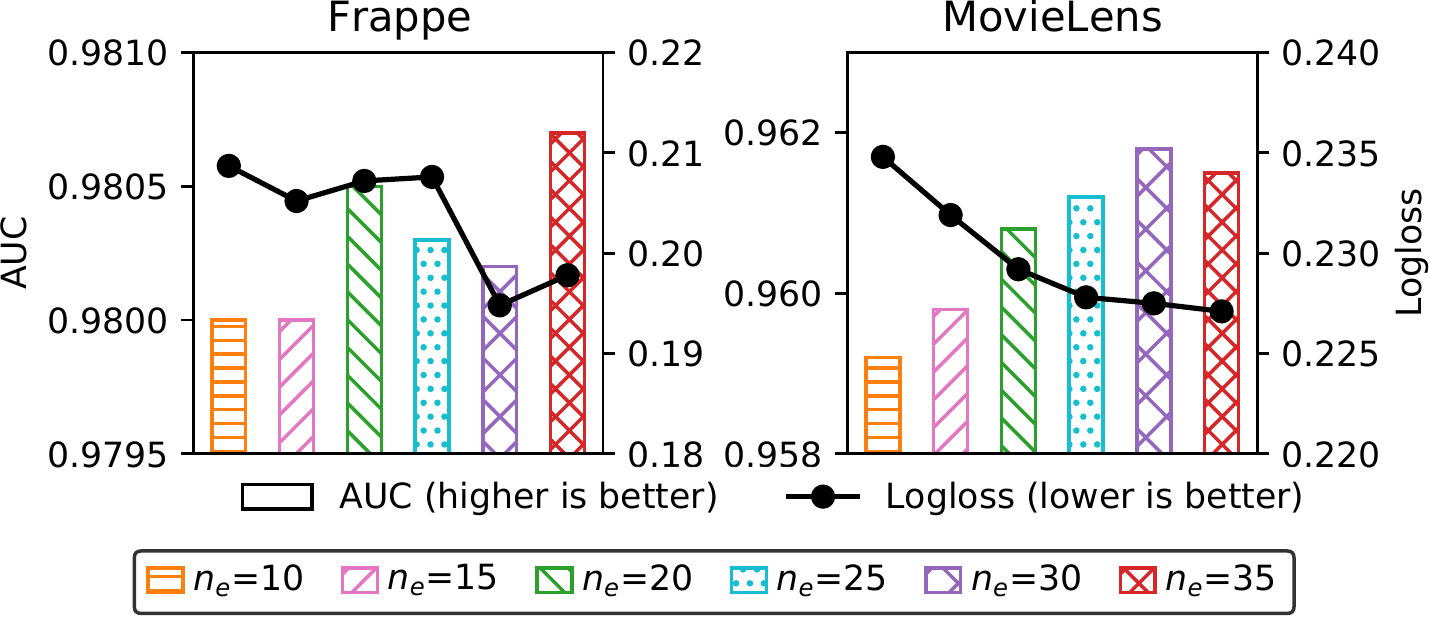}
    \caption{\update{The impact of increasing the input embedding size on the prediction performance of \nameplus.}}\vspace{-3mm}
    \label{fig:nemb}\vspace{-4mm}
\end{figure}

\vspace{1mm}
\noindent
\powerpoint{\update{Enhancing \nameplus with a Larger Embedding Size.}}
For a fair comparison, all the results reported in Table~\ref{tab:overall_resutls} consistently adopt an embedding size of 10.
To evaluate the impact of the embedding size $n_e$ on the prediction performance of \nameplus, we adopt the best \nameplus and show the results of varying $n_e$ in Figure~\ref{fig:nemb} measured in AUC and $\logloss$.
We can find that increasing the embedding size can further improve the prediction performance.
Particularly, the AUC increases from 0.9800 to 0.9807 on \frappe and from 0.9592 to 0.9615 on \movielens when adopting a larger embedding size of 35.
The results suggest that by embedding the input into a larger latent space, we can further increase the model capacity of \nameplus.

\begin{table}[t]
    \small
    \centering
    \caption{Top Global Interaction Terms for \diabetes.}
    \label{tab:diabetes_interaction_terms}
    \begin{tabular}{cc|c}
    \toprule[1.8pt]
    
    Frequency     &   Orders   &   Interaction Term \\ \midrule[1pt]
    3.75     &   1     &   (inpatient\_score)      \\
    2.41     &   1     &   (diag\_1\_category)      \\
    2.39     &   2     &   (A1Cresult, glimepiride)      \\
    2.10     &   2     &   (nateglinide, glyburide\_metformin)      \\
    1.87     &   1     &   (num\_diagnoses)      \\
    1.65     &   3     &   (metformin, nateglinide, glyburide\_metformin)      \\
    1.45     &   2     &   (num\_diagnoses, diabetes\_med)      \\
    1.36     &   2     &   (inpatient\_score, diabetes\_med)      \\
    \bottomrule[1.8pt]
    \end{tabular}\vspace{-2mm}
\end{table} 

\subsection{Interpretability}
\label{sec:exp_interpretability}

\vspace{1mm}
\noindent
\powerpoint{\update{Interpretability Settings.}}
We demonstrate the interpretability results of \framework in two representative domains, namely the app usage prediction on \frappe and the diabetes readmission prediction on \diabetes.
The learning task on \frappe is to predict the app usage status given the use context.
The context includes 10 attribute fields, \textit{\{user\_id, item\_id, daytime, weekday, weekend, location, is\_free, weather, country, city\}}, which together characterize the usage patterns of mobile end-users;
For \diabetes, the learning task is to predict the probability of readmission by analyzing factors related to readmission and other outcomes of patients with diabetes.
There are 43 attribute fields for the prediction, and we show the 10 most important ones for illustration.
The explanations of the attribute fields for both datasets are publicly available~\cite{frappe15,diabetes}, with which the interpretability results produced by \framework can be validated.

For both datasets, we first show the \textit{global feature importance} of respective attribute fields obtained by aggregating the value vectors of exponential neurons as discussed in Section~\ref{sec:analysis_and_discussion}, and compare the global feature attribution of \name with two widely adopted interpretation approaches Lime~\cite{lime} and Shap~\cite{shap}, which are input perturbation interpretation methods based on linear regression and game theory respectively to identify the feature importance of the models to be interpreted.
Specifically, the interpretation results of Lime and Shap for \frappe and \diabetes is based on the best single-model baseline DNN and GAT~\cite{gat} respectively, and the global feature importance given by both methods are obtained by the aggregation of local feature attribution of all instances of the test dataset.
We then show the top interaction terms captured by \name with corresponding \textit{frequency} and \textit{orders}, which denote the average occurrence count per instance\footnote{Note that there are $K \cdot o$ exponential neurons in \name, and each neuron captures a specific interaction term given the input instance dynamically.} and the number of features captured for each interaction term respectively.
We also illustrate local interpretation by showing feature interaction weights assigned by \mname-Module via aggregation, and again compare the local feature attribution results of \name with Lime and Shap.

\begin{figure}[t!]
  \centering
  \begin{minipage}{.44\linewidth}
    \centering
      {\includegraphics[clip, trim=14cm 0.0cm 0.0cm 0.52cm,width=1.0\linewidth]{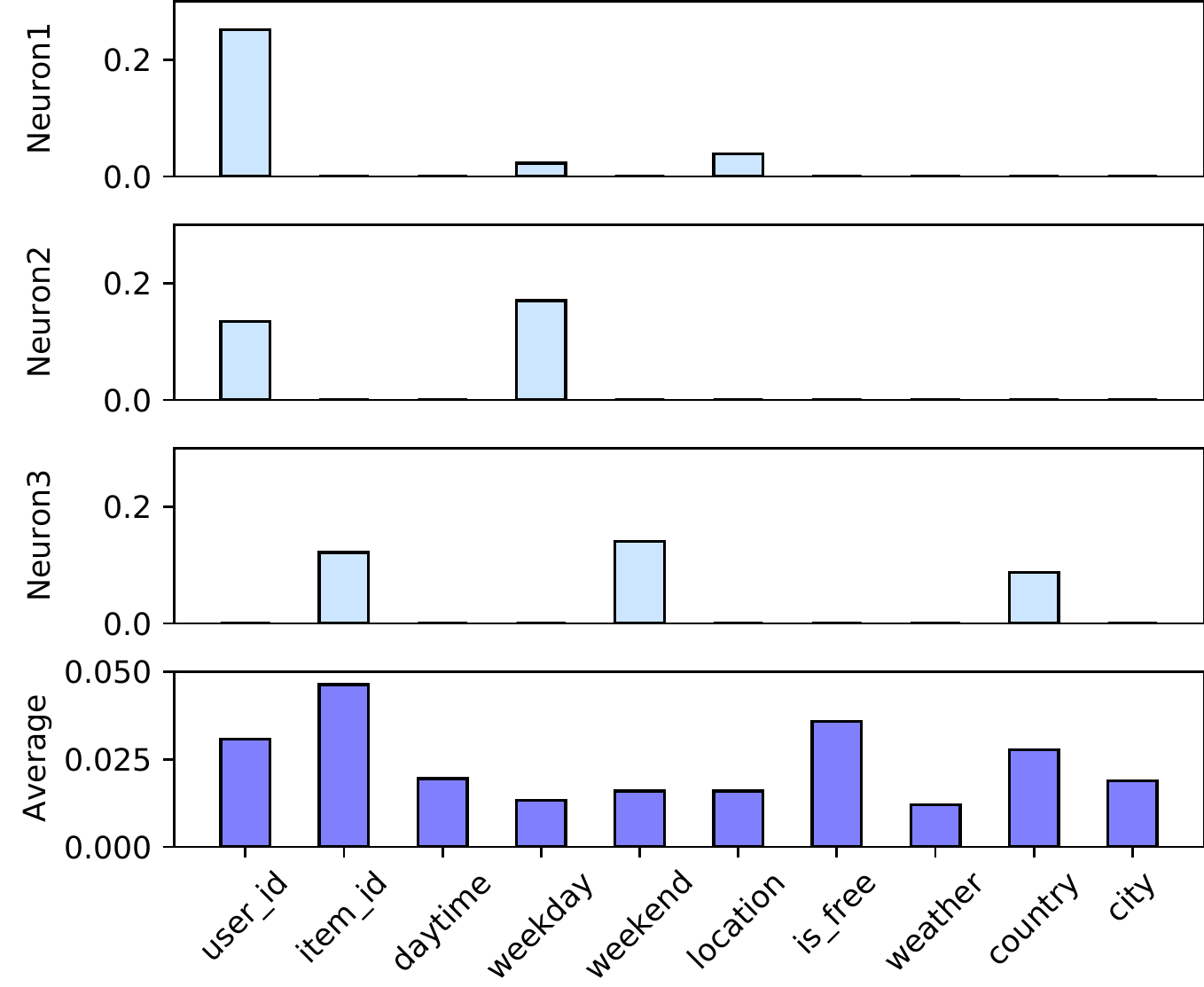}
      }
    
  \end{minipage}\quad
  \begin{minipage}{.5\linewidth}
    \centering
      {\includegraphics[width=1.16\linewidth]{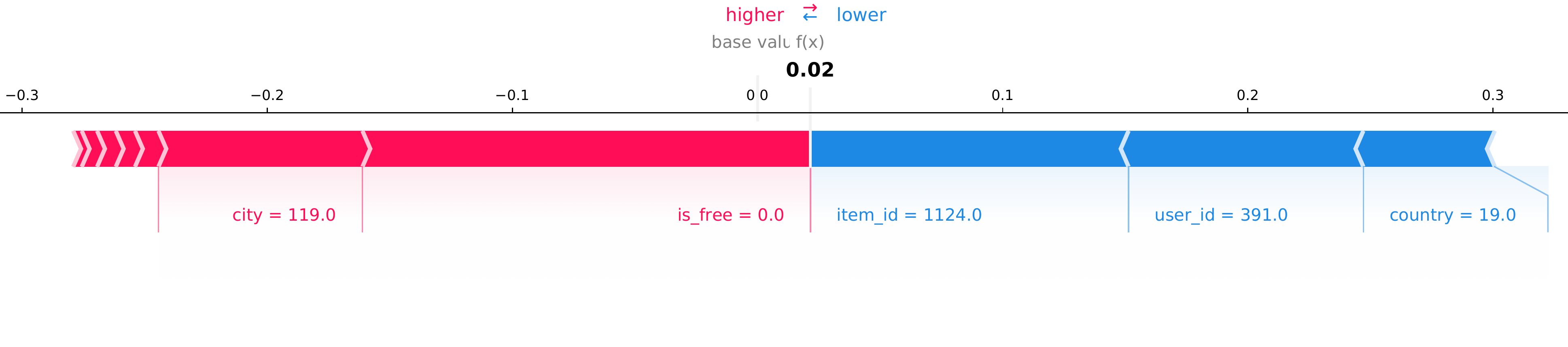}}

      {\includegraphics[width=1.16\linewidth]{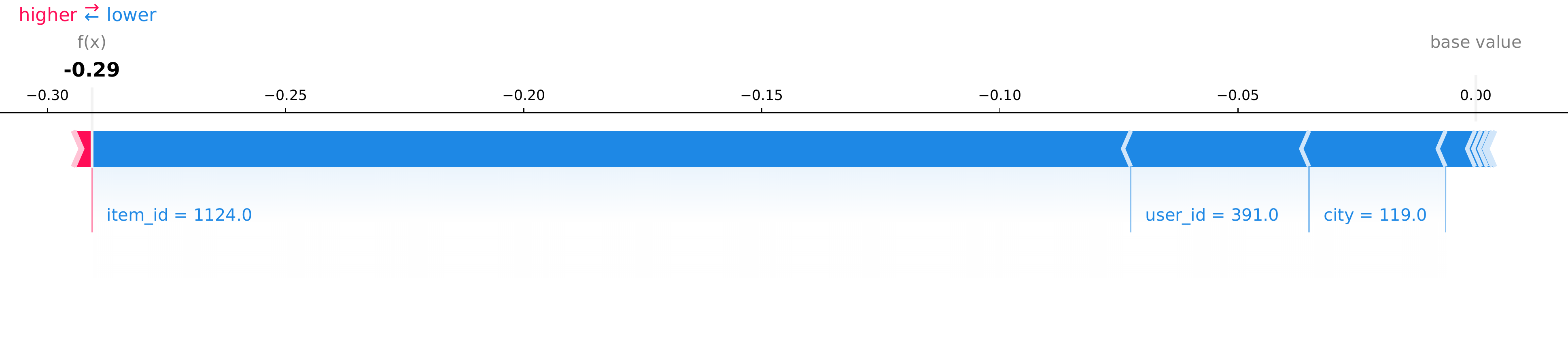}}

  \end{minipage}
  \vspace{-5mm}
  \caption{\update{Local feature attribution with \name (left) and local feature importance weights given by Lime (right top) and Shap (right bottom) on \frappe.}\label{fig:frappe_local}}
  \vspace{-3mm}
\end{figure}

\vspace{1mm}
\noindent
\powerpoint{\update{Global Interpretability.}}
We illustrate the \textit{global feature attribution} in Figure~\ref{fig:global_frappe_diabetes} and summarize the top interaction terms captured by \name in Table~\ref{tab:frappe_interaction_terms} and Table~\ref{tab:diabetes_interaction_terms} for the two datasets respectively.

From Figure~\ref{fig:global_frappe_diabetes}, we can observe that the most important features identified by \name for \frappe are \textit{\{user\_id, item\_id, is\_free\}}.
The global focus on these attributes are reasonable since \textit{user\_id} and \textit{item\_id} identify the user and item and are the two primary features used in learning tasks such as collaborative filtering, and \textit{is\_free} indicates whether the user pays for the app, which is highly correlated with the preference of the user for the app~\cite{frappe15}.
Likewise, the features of the highest importance identified by \name for \diabetes include \textit{\{emergency\_score, inpatient\_score, num\_diagnoses\}}, which is in line with the coefficients of attribute fields estimated by the logistic regression model in~\cite{diabetes}.
We can also notice that the global feature importance provided by \name is consistent with the two general interpretation approaches, namely Lime and Shap.
However, we note that the interpretation results provided by \name are relatively more reliable since \name inherently supports global feature attribution and its modeling process is more transparent, whereas Lime and Shap are typically adopted as a medium to interpret other ``black box'' models via approximation.


\begin{figure}[t!]
  \centering
  \begin{minipage}{.44\linewidth}
    \centering
      {\includegraphics[clip, trim=14cm 0.0cm 0.0cm 0.52cm,width=1.0\linewidth]{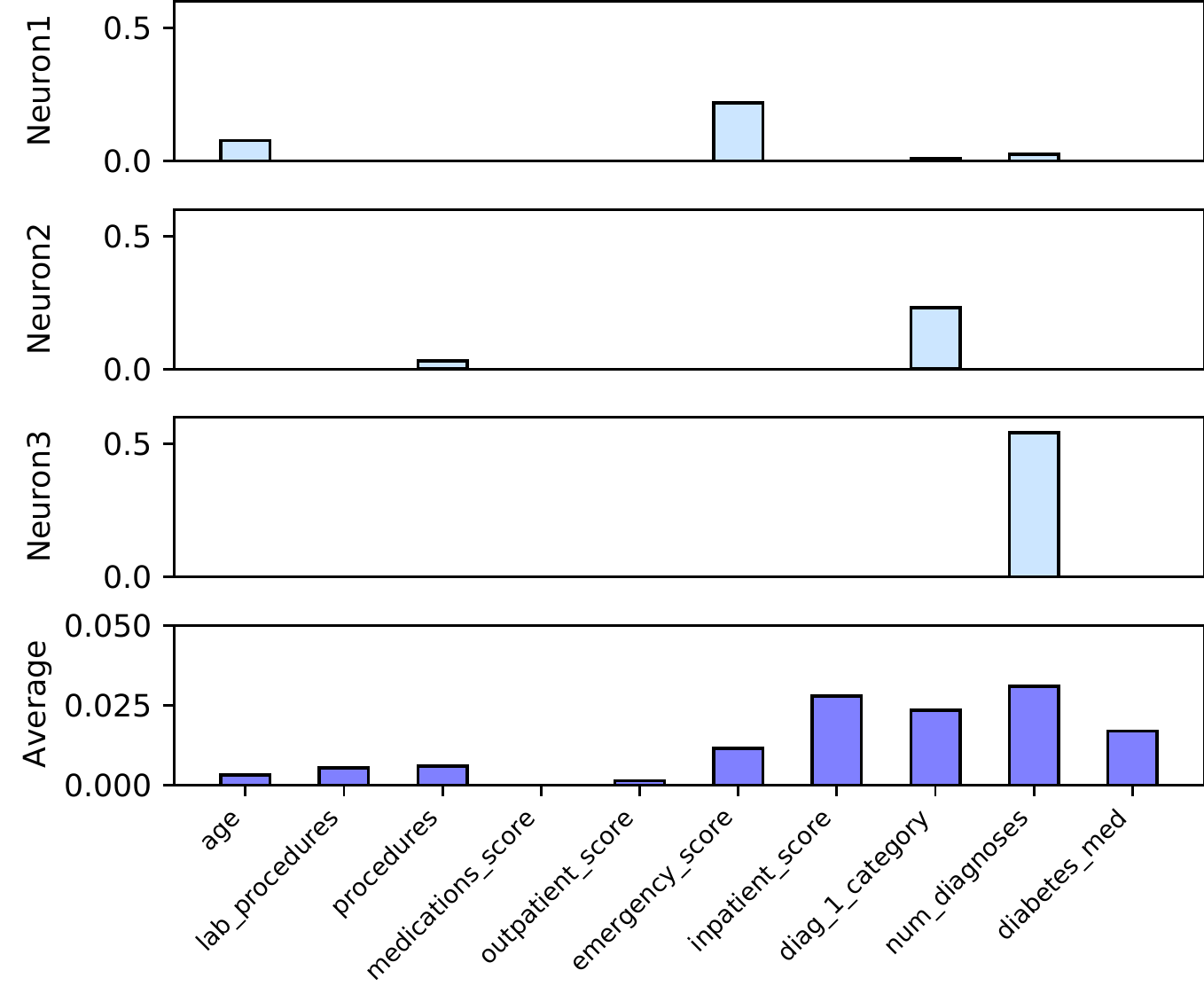}}
    
  \end{minipage}\quad
  \begin{minipage}{.5\linewidth}
    \centering
      {\includegraphics[width=1.16\linewidth]{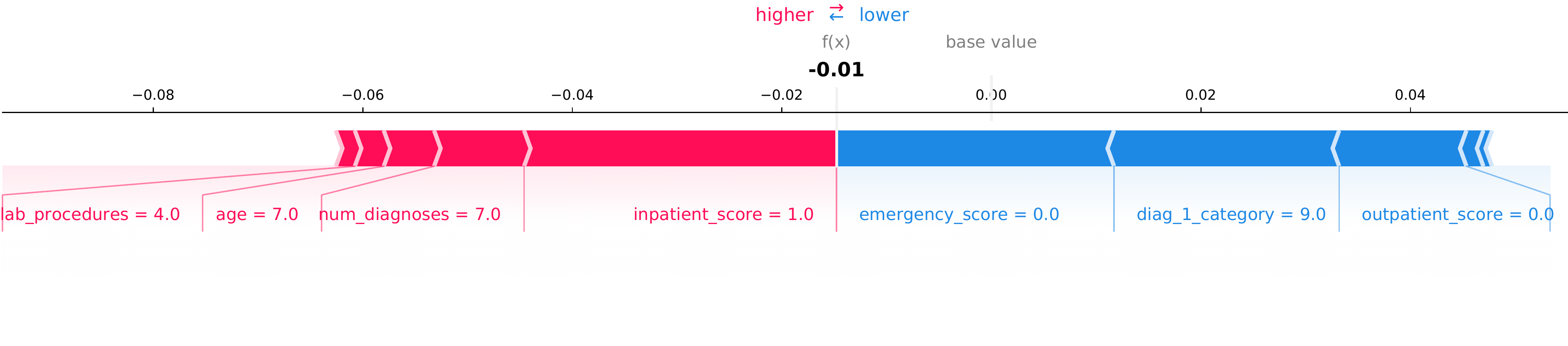}}

      {\includegraphics[width=1.16\linewidth]{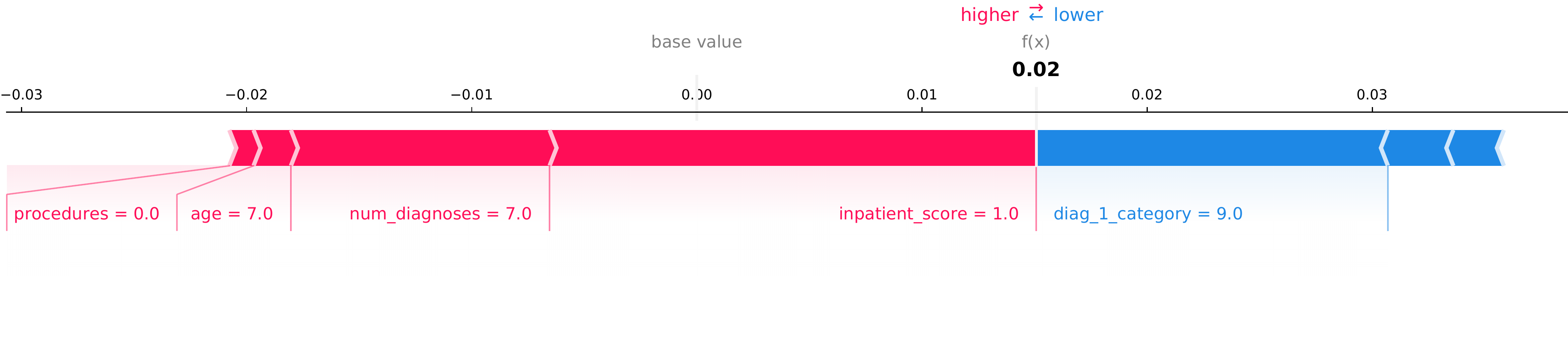}}

  \end{minipage}
  \vspace{-0mm}
  \caption{\update{Local feature attribution with \name (left) and local feature importance weights given by Lime (right top) and Shap (right bottom) on \diabetes.}\label{fig:diabetes_local}}
  \vspace{-6mm}
\end{figure}

From the top global interaction terms for \frappe in Table~\ref{tab:frappe_interaction_terms}, we can find that firstly, the most frequently modeled attribute fields for the interaction terms include \textit{use\_id}, \textit{item\_id} and \textit{is\_free}, which is consistent with the global feature importance in Figure~\ref{fig:global_frappe_diabetes}.
Secondly, these interaction terms occur quite frequently in the interaction modeling, e.g., the frequency of the interaction term \textit{(weekday, location, is\_free)}, \textit{(item\_id, is\_free, city)} and \textit{(item\_id, is\_free)} are 3.71, 3.36 and 3.22 respectively, which indicates that these cross features (with different interaction weights) are used multiple times on average for each instance (note that there are $K \cdot o$ interaction terms during inference).
Thirdly, the orders of the top interaction terms are mostly 2 and 3, which suggests that identifying the proper set of attributes for interaction modeling is necessary, and capturing cross features by enumerating all possible feature combinations is inefficient and ineffective, which may simply introduce noise.

From the top global interaction terms for \diabetes in Table~\ref{tab:diabetes_interaction_terms}, we can observe that the most frequently modeled attribute fields in the interaction terms are quite diversified, indicating that different exponential neurons indeed capture diverse cross features, which is more parameter efficient in modeling feature interactions. 
Further, the orders of the top interaction terms are less than 3, and there are many first-order terms, which indicates that for some datasets such as \diabetes, modeling cross features of high interaction orders may not be necessary.

\vspace{1mm}
\noindent
\powerpoint{\update{Local Interpretability.}}
The local feature attribution of \name for a representative input instance of \frappe is illustrated in Figure~\ref{fig:frappe_local}, which shows the interaction weights of three representative exponential neurons and the average weights of all neurons.
We can notice that different exponential neurons capture distinct cross features selectively in a sparse manner.
For example, Neuron3 captures feature interaction term \textit{(item\_id, weekend, country)}, which indicates that for this specific instance, Neuron3 is responsive to these three attributes.
Further, the aggregated interaction weights for this instance show that \textit{item\_id}, \textit{is\_free} and \textit{user\_id} are the three most discriminative attributes, which is consistent with the global interpretation results in Figure~\ref{fig:global_frappe_diabetes}.
We also illustrate the local feature attribution by Lime~\cite{lime} and Shap~\cite{shap} in Figure~\ref{fig:frappe_local}.
We can notice that although both Lime and Shap identify \textit{item\_id}, \textit{user\_id} and \textit{city} as the three most important features as \name, Lime also assigns large importance weight to other features, namely \textit{is\_free} and \textit{country}, which indicates that external interpretation approaches may not be consistent and reliable as they are only approximating the models to be interpreted from different perspectives.

Similar local feature attribution results for \diabetes can be found in Figure~\ref{fig:diabetes_local}.
We can find that different exponential neurons lay emphasis on distinct cross features.
Specifically, Neuron1 and Neuron2 are more attentive to \textit{emergency\_score} and \textit{diag\_1\_category} respectively, and Neuron3 are more focused on \textit{num\_diagnoses}.
Further, for this specific diabetes patient, the last five features, i.e., \textit{emergency\_score}, \textit{inpatient\_score}, \textit{diag\_1\_category}, \textit{num\_diagnoses} and \textit{diabetes\_med} are the most informative attributes for the readmission prediction.
With such local interpretation, \framework can support more personalized analytics and management.


\section{Related Work}
\label{sec:related work}

\vspace{1mm}
\noindent
\highlight{Feature Interaction Modeling.}
Cross features explicitly model feature interactions among attribute fields by multiplying corresponding constituent features, which proves to be essential for the predictive analytics of various applications, e.g., app recommendation~\cite{wideanddeep} and click prediction~\cite{shan2016deep}.
Many existing works~\cite{wideanddeep, shan2016deep} capture cross features in an implicit manner with DNNs.
However, modeling multiplicative interaction implicitly with DNNs requires a substantial number of hidden units~\cite{latentcross,polynomialnn,multiplicative}, which makes the modeling process inefficient and less interpretable~\cite{afn,tracer,xaisurvey}.

Alternatively, many models~\cite{fm, ffm, afm, hofm, deepandcross, pnn, xdeepfm} propose to capture cross features explicitly, which generally obtains better prediction performance.
Among these studies, ~\cite{fm, ffm, afm} capture second-order feature interactions, and ~\cite{hofm} models higher order interactions feature interactions within a predefined maximum order.
Many explicit interaction modeling studies further integrate a DNN to enhance the modeling capacity as DNNs are powerful at capturing fine-grained non-linear interactions.
Representative studies include: (i) NFM~\cite{nfm} with bi-interaction pooling, i.e., an element-wise product between input feature embeddings, (ii) DeepFM~\cite{deepfm} as an ensemble model of FM and DNN, (iii) DCN and its DNN ensemble model with feature cross operations on input feature embeddings, (iv) PNN~\cite{pnn} with both inner and outer pairwise products of input feature embeddings, and (v) CIN and its DNN ensemble model xDeepFM~\cite{xdeepfm} with compressed interaction of input feature embeddings.
Recently, AFN~\cite{afn} propose to model cross features of arbitrary orders with logarithmic neurons, which however has the inherent limitation of only positive input due to logarithmic transformation and meanwhile lacks runtime flexibility.
\name instead models feature interactions with exponential neurons adaptively with gated attention, which is more effective, interpretable and parameter-efficient.


\vspace{1mm}
\noindent
\highlight{\update{Interpretability.}}
With the increasingly dominant role played by machine learning models,
there is a growing demand for model transparency and interpretability~\cite{xaisurvey,gade2019explainable}, which can help debug the learning models and contribute to the verification and improvement of these models.
In addition, an interpretable model also improves human understanding of many domains
and engenders trust in the analytical results~\cite{molnar, gade2019explainable}.

A simple yet effective approach acting as either global or local interpretability is feature attribution, which produces the feature importance for input instance in terms of the weight and magnitudes of the features being used~\cite{xaisurvey,gade2019explainable,lime,shapley1953value}, e.g., attributing a medical diagnosis to each lab test input.
There exist general interpretation methods in this line of research.
Notably, Shapley value~\cite{shapley1953value,shap} assesses the importance of each feature in model predictions based on game theory.
LIME~\cite{lime} builds a linear model to locally approximate the model by input perturbation, which provides model-agnostic local explanations.
Grad-CAM~\cite{selvaraju2017grad} provides a visual explanation based on gradient-weighted class activation mapping for CNN-based models to highlight local regions.

There are also domain-specific interpretation methods with integrated domain expertise.
For instance, in healthcare analytics and finance, deep models are increasingly being adopted to achieve high predictive performance~\cite{purushotham2018benchmarking,Zheng2017bias,Zheng2017irregularity,Zheng2021pace}; however, such critical and high-stakes applications stress the need for interpretability.
Particularly, the attention mechanism~\cite{bahdanau2014neural} is widely adopted to facilitate the interpretability of deep models by visualizing the attention weights. 
With the attention mechanism integrated into the model design, many studies~\cite{ma2017dipole,choi2016retain,tracer}
manage to achieve interpretable healthcare analytics.
Specifically, Dipole~\cite{ma2017dipole} supports the visit-level interpretation in the diagnosis prediction with three attention mechanisms. 
RETAIN~\cite{choi2016retain} and TRACER~\cite{tracer} can support both the visit-level and the feature-level interpretation.
However, one inherent limitation of most existing methods is that their interpretability is built upon single input features, while disregarding feature interactions that are essential for relational analytics.

\section{Conclusion}
\label{sec:conclusion}

Existing DNNs, while producing super-human results for certain application domains, have not been shown to produce meaningful results when applied to structured data.
To fill this gap, we present a learning model tailored for structured data, called \name, and a lightweight framework \framework based on \name for relational data analytics, \update{which is accurate, efficient and more interpretable. }
The key idea is to model attribute dependencies and correlations selectively and dynamically, through cross features, derived by first transforming input features into exponential space, and then determining interaction weights and the interaction order adaptively for each cross feature. 
To model cross features of arbitrary orders dynamically and filter noisy features selectively, we propose a novel gated attention mechanism to generate interaction weights given the input tuple.
Consequently, \name can identify the most informative cross features in an input-aware manner for more accurate prediction and better interpretability during inference.
An extensive experimental study on real-world datasets confirms that 
our \name consistently achieves superior prediction performance compared to existing models, \update{and \framework facilitates global interpretability and local instance-aware interpretability.}

\section{ACKNOWLEDGMENTS}\label{sec:acknowledgement}

We thank the anonymous reviewers for their constructive comments.
This research is supported by Singapore Ministry of Education Academic Research Fund Tier 3 under MOE's official grant number MOE2017-T3-1-007. 
H. V. Jagadish was supported in part by NSF grants 1741022 and 1934565.
Meihui Zhang's work is supported by National Natural Science Foundation of China (62050099).

\newpage
\bibliographystyle{ACM-Reference-Format}
\bibliography{reference}

\end{document}